\documentclass[letterpaper]{article} 
\PassOptionsToPackage{table}{xcolor}
\usepackage{aaai2027}  

\usepackage{times}
\usepackage{helvet}
\usepackage{courier}
\usepackage[hyphens]{url}
\usepackage{graphicx}
\usepackage{microtype}
\usepackage{natbib}
\usepackage{caption}
\usepackage{amsmath,amssymb}
\usepackage{booktabs}
\usepackage{multirow}
\usepackage{array}
\usepackage[switch]{lineno}

\title{Degradation-Guided Underwater Image Restoration with Task-Oriented \\Latent Control}

\author{
    Xu Zhang\textsuperscript{\rm 1},
    Xuhui Cao\textsuperscript{\rm 1},
    Kangzhe Yuan\textsuperscript{\rm 1},
    Laibin Chang\textsuperscript{\rm 2},
    Yichu Xu\textsuperscript{\rm 3},
    Shi Chen\textsuperscript{\rm 4},
    Huan Zhang\textsuperscript{\rm 5},
    Yong Chen\textsuperscript{\rm 1}\thanks{Corresponding author.}
}

\affiliations{
    \textsuperscript{\rm 1}National Engineering Research Center for Multimedia Software, School of Computer Science, Wuhan University\\
    \textsuperscript{\rm 2}School of Computer and Artificial Intelligence, Zhengzhou University\\
    \textsuperscript{\rm 3}Aerospace Information Research Institute, Henan Academy of Sciences\\
    \textsuperscript{\rm 4}Department of Computer Science, University of Macau\\
    \textsuperscript{\rm 5}School of Information Engineering, Guangdong University of Technology\\
    \{zhangx0802, xuhuicao, 2025305233105, changlb666, xuyichu, sggychen\}@whu.edu.cn,
chenshi@um.edu.mo, huanzhang2021@gdut.edu.cn
}

\begin{document}
\maketitle
\begin{abstract}
Degradation information in underwater images plays a dual role:
its spatial and spectral cues can guide adaptive restoration, while
degradation-entangled features may be propagated without explicit regulation
during decoding.
Existing methods largely overlook this dual role, either underexploiting degradation cues or directly forwarding encoder features through skip connections. To address this issue, we propose PROTEUS, which couples degradation-guided feature adaptation with task-oriented latent control. PROTEUS tackles this problem from two complementary perspectives. At the feature level, the Guided Dynamic Feature Modulation Block exploits spatially varying degradation cues to adapt feature processing across network stages. At the representation level, the task-oriented latent controller learns a structured control code under discriminative regularisation and uses it for channel-wise modulation of skip features, without requiring the code to form a metrically cleaner embedding. Extensive experiments on five paired and four non-reference underwater benchmarks demonstrate that PROTEUS achieves highly competitive restoration performance, with a favourable balance between restoration quality and computational cost.
\end{abstract}

\section{Introduction}

Underwater imagery supports a wide range of marine applications, yet the aquatic medium introduces degradations far more complex than their terrestrial counterparts. Wavelength-dependent absorption, backscatter haze, and spatially varying illumination co-occur and interact, producing entangled colour casts, reduced contrast, and depth-dependent intensity attenuation that directly impair downstream tasks such as object detection \cite{RRNet,Nested,clb_detection}, segmentation \cite{seg_1,seg_2}, and 3-D reconstruction \cite{3D_1, 3D_2}.

Classical model-based methods invert a simplified image formation model~\cite{jaffe1990computer} by estimating the transmission map and background light. Although physically interpretable, they rely on hand-crafted priors such as the dark channel prior~\cite{he2010single,UDCP} that are easily violated in practice, leading to over-compensation or residual colour shifts.

\begin{figure}[!t]
    \centering
    \includegraphics[width=\columnwidth]{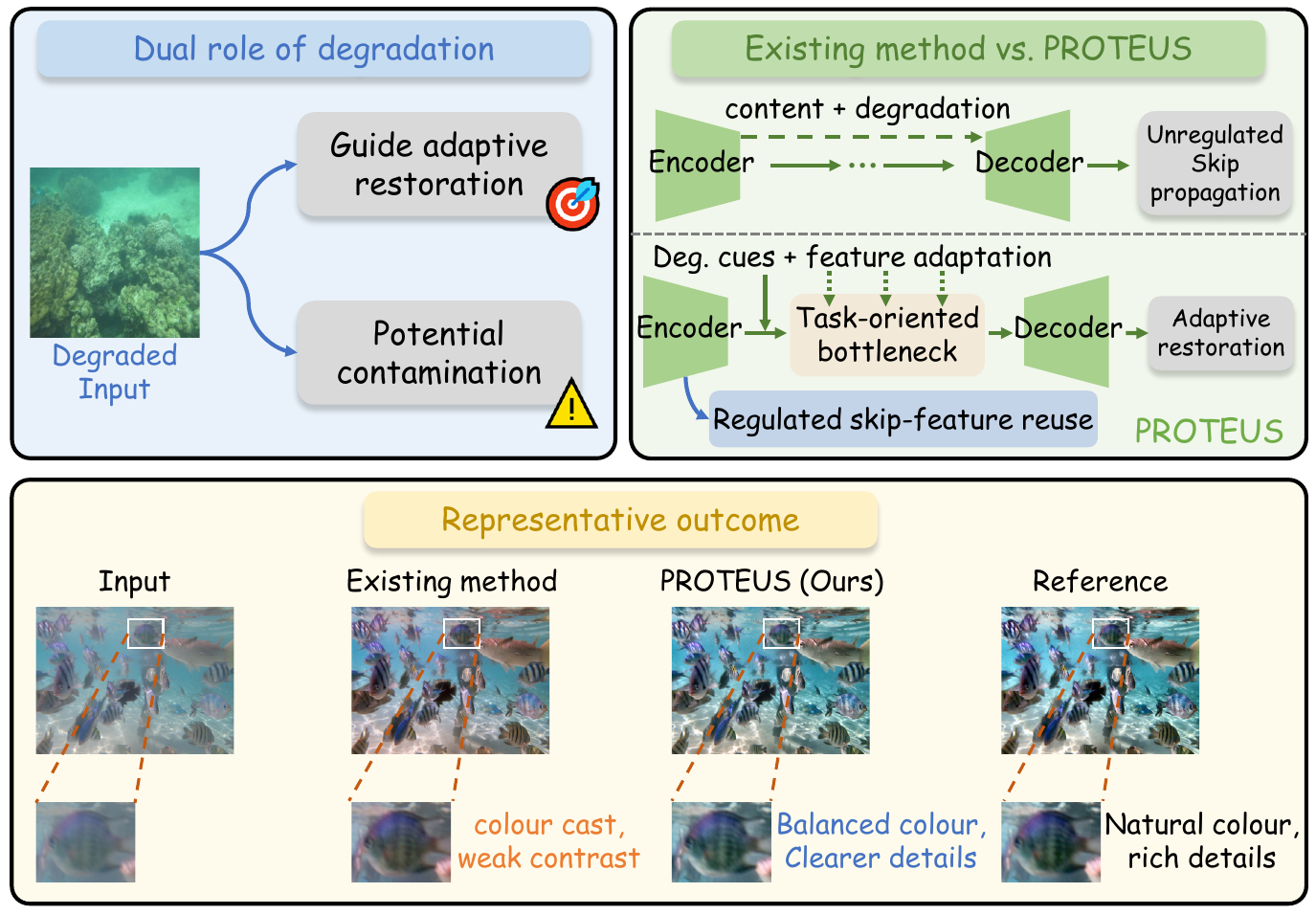}
\caption{Motivation and overview of PROTEUS.
(a) Degradation information provides useful restoration cues but may also
remain entangled in encoder features.
(b) Unlike conventional U-Net pipelines that directly propagate encoder
features through skip connections, PROTEUS combines degradation-guided feature
adaptation with task-oriented latent control to regulate skip-feature reuse.
(c) Representative comparison on U90, where PROTEUS produces more
balanced colour and clearer local details.}
    \label{fig:motivation}
    \vspace{-1em}
\end{figure}

Learning-based methods \cite{UIEB, Ucolor, Ushape, NU2Net, Semi-UIR, FUnIE, GUPDM, zhou_IJCV_ADPP, zhou_IJCV_HCLR, WWVP, FSMC,zhou_TCSVT_FDCE, zhang_TCSVT_PCFB, zzk_PR, UniUIR} achieve impressive gains by training deep networks in a data-driven fashion, yet often
reuse encoder features through skip connections without explicitly regulating
the degradation information embedded in them. This reveals a dual role of
underwater degradation information: its spatial and spectral patterns provide
useful cues for adaptive restoration, while degradation factors may also remain
entangled with encoder representations and be propagated to the decoder.

As illustrated in Fig.~\ref{fig:motivation}(a)--(b), PROTEUS addresses these
two aspects jointly. It exploits spatially varying degradation cues to guide
stage-wise feature adaptation, while a task-oriented bottleneck code regulates
the reuse of encoder features during decoding.

At the feature level, the Guided Dynamic Feature Modulation Block (GDFMB) exploits spatially varying degradation cues from the input image to guide stage-wise feature modulation, enabling adaptive correction of non-uniform degradations throughout the encoder-decoder hierarchy. At the representation level, the task-oriented latent controller structures a bottleneck code and uses it to regulate skip-feature channels during reconstruction. This operational role does not require the code to be closer to a clean-image latent under raw feature-space distance. In addition, a Gradient Fusion Block enhances detail preservation during feature transformation, while a Gray-Edge Prior Module provides a colour-bias mitigation strategy before feature extraction.

The main contributions are summarized as follows:
\begin{itemize}
    \item We formulate underwater restoration around the dual role of degradation
information: exploiting useful cues for adaptive processing while regulating
degradation-entangled feature reuse.
    \item We introduce a task-oriented latent controller that learns a structured control code under discriminative regularisation and uses it for channel-wise skip modulation.
    \item We design the GDFMB, which leverages spatially varying degradation cues to guide stage-wise feature modulation, enabling adaptive correction of non-uniform underwater distortions across the encoder-decoder network.
\end{itemize}

\section{Related work}
\subsection{Underwater Image Restoration}

Underwater image restoration methods can be broadly divided into model-based, handcrafted, and learning-based approaches. Model-based methods recover scene radiance by estimating physical parameters under the Jaffe--McGlamery model~\cite{jaffe1990computer}, as exemplified by UDCP~\cite{UDCP} and IBLA~\cite{IBLA}. Handcrafted methods, including CLAHE~\cite{zuiderveld1994contrast}, white-balance fusion~\cite{ancuti2012enhancing}, and Retinex-based enhancement~\cite{zhang2022underwater}, avoid explicit physical modelling but remain sensitive to scene-dependent degradation due to their fixed priors.

Learning-based methods have become the dominant paradigm by learning restoration mappings directly from paired or unpaired data~\cite{UIEB,Ucolor,Ushape,NU2Net,Semi-UIR,GUPDM,zhou_IJCV_HCLR,UniUIR}. Representative approaches incorporate confidence-guided multi-input fusion~\cite{UIEB}, transmission-aware multi-colour-space features~\cite{Ucolor}, channel-wise multi-scale attention~\cite{Ushape}, semi-supervised learning~\cite{Semi-UIR}, contrastive learning~\cite{jiang2024five}, or uncertainty-guided fusion~\cite{fu2022uncertainty}. However, encoder features are still commonly reused through skip connections during decoding without explicit latent regulation. 
Our method instead learns a task-oriented bottleneck code to regulate
skip-feature reuse, without assuming a clean--degradation decomposition.

\subsection{Degradation-Adaptive Feature Processing}

Recent restoration methods improve input-adaptive processing mainly through
content-adaptive feature interaction and degradation-conditioned transformation.
Transformer-based backbones such as SwinIR~\cite{liang2021swinir},
Restormer~\cite{Restormer}, DAT~\cite{chen2023dual}, and
DRSformer~\cite{DRSformer} primarily adapt \emph{where} and \emph{how}
information is aggregated according to image content. In parallel,
degradation-aware methods such as AirNet~\cite{AirNet},
PromptIR~\cite{PromptIR}, OneRestore~\cite{OneRestore}, and
MPerceiver~\cite{MPerceiver} derive degradation representations, prompts, or
control signals to condition the restoration process.

Despite their effectiveness, content-adaptive interaction and explicit
degradation conditioning are not always jointly modelled. Generic attention
mechanisms do not necessarily convert degradation cues into targeted feature
transformation, whereas many conditioning methods compress degradation
information into global embeddings or coarse prompts. Such representations may
be insufficient for underwater degradations that vary jointly across spatial
regions and colour channels. Our GDFMB therefore extracts spatially varying
degradation cues from the input and uses them to guide feature transformation
across network stages.

\subsection{Latent Representation Learning and Control}

Latent representation learning has been explored in image restoration to
separate task-relevant information from degradation factors. BaryIR~\cite{BaryIR}
maps degraded representations towards a clean manifold through barycentric
projection and encourages orthogonality between clean and degradation
components. Contrastive methods such as AECR-Net~\cite{wu2021contrastive} and
Semi-UIR~\cite{Semi-UIR} further improve restoration representations by
constructing positive and negative relations at the image or feature level.

These methods mainly regularise latent geometry or improve degradation
discrimination. Existing degradation-aware restoration methods typically use
learned representations or prompts to condition feature transformation
\cite{AirNet,PromptIR,OneRestore,MPerceiver}, while their role in regulating
encoder-feature reuse during decoding remains underexplored. PROTEUS instead
turns the bottleneck representation into an explicit control signal for
skip-feature reuse.

\begin{figure*}[!htb]
	\centering
	\includegraphics[width=\linewidth]{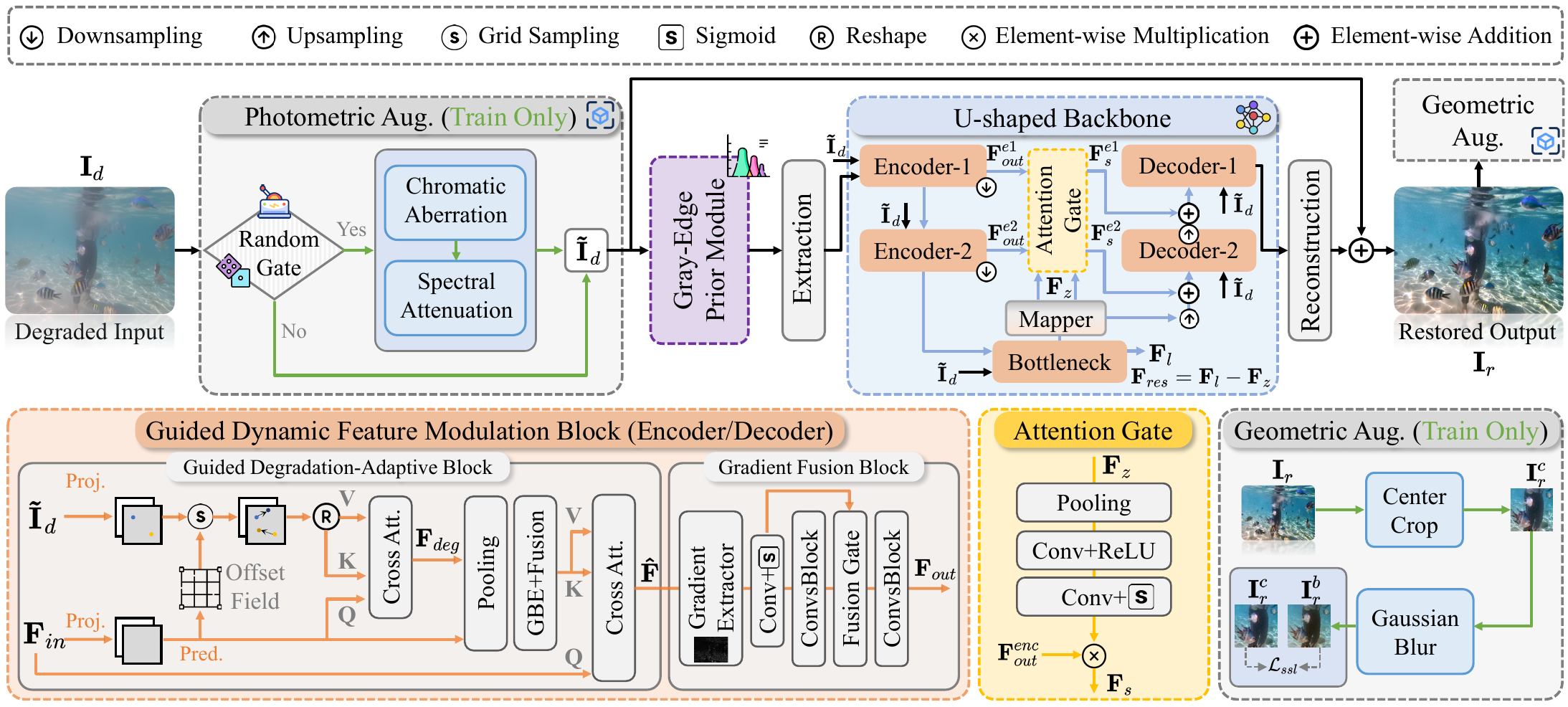}
\caption{Proposed network architecture of PROTEUS. The framework combines
degradation-guided feature processing and task-oriented latent control within
a U-shaped backbone. A guide image generated by the Gray-Edge Prior Module
conditions encoder and decoder features across multiple stages, while a
structured bottleneck code modulates skip features before their fusion into
the decoder. The global residual path facilitates stable reconstruction, and
the photometric and geometric augmentation branches are used only during
training to improve robustness to diverse underwater degradations.}
	\label{fig:model}
    \vspace{-1em}
\end{figure*}

\section{Proposed Method}\label{sec:method}

\subsection{Overview}

Given a degraded underwater image
$\mathbf{I}_{{d}} \in \mathbb{R}^{3 \times H \times W}$, our
goal is to produce a restored image
$\mathbf{I}_{r} \in \mathbb{R}^{3 \times H \times W}$ that
faithfully recovers the scene's colour, contrast, and structural
details. As illustrated in Fig.~\ref{fig:model}, the overall
forward pass follows a global residual formulation:
\begin{equation}\label{eq:overall}
    {\mathbf{I}_{r}} = \mathcal{D}\!\bigl(
        \mathcal{P}(\mathcal{E}(\mathbf{I}_{{d}};\,
        \mathbf{\tilde{I}}_{d}))\,;\;
        \mathbf{\tilde{I}}_{d}
    \bigr) + \mathbf{I}_{{d}},
\end{equation}
where $\mathcal{E}$ denotes the encoder, $\mathcal{P}$ the
bottleneck latent controller, $\mathcal{D}$ the decoder,
and $\mathbf{\tilde{I}}_{d}$ the guide image. The global residual connection ensures that
the network only needs to learn the restoration residual.

The backbone is a symmetric three-level U-Net. A $3\times3$ convolutional stem maps the input to base feature dimension $C=32$. The encoder processes features at resolutions $H\times W$ and $\frac{H}{2}\times\frac{W}{2}$ before reaching a bottleneck at $\frac{H}{4}\times\frac{W}{4}$. At every encoder and decoder stage, a Guided Dynamic Feature Modulation Block comprising a Guided Degradation-Adaptive Block followed by
a Gradient Fusion Block with a residual blend ratio of 0.5 serves as the core processing unit. Downsampling and upsampling are handled by PixelUnshuffle and PixelShuffle of factor 2, each paired with a $3\times3$ convolution for channel adjustment. At the bottleneck, the task-oriented latent controller $\mathcal{P}$ maps the deepest feature map to a structured control code and a complementary residual; the code is decoded directly and also modulates skip connections through channel-wise Attention Gates. A final $3\times3$ convolution produces the three-channel output.

\subsection{Photometric and Geometric Augmentation}

To improve robustness under diverse underwater conditions, we apply
two physics-motivated transformations to the input image during
training. A {chromatic aberration} step shifts the red and blue
channels by random offsets while keeping green fixed, simulating
wavelength-dependent refraction:
\begin{equation}
  \mathbf{I}'_{d} = \bigl[\,\mathcal{S}(\mathbf{I}_{d}^R,\,\boldsymbol{\delta}_R),\;
  \mathbf{I}_{d}^G,\;
  \mathcal{S}(\mathbf{I}_{d}^B,\,\boldsymbol{\delta}_B)\bigr],
  \quad \boldsymbol{\delta}_{\{R,B\}} \sim \mathcal{U}(-s,\,s)^{2},
\end{equation}
where \(\mathcal{S}(\cdot,\boldsymbol{\delta})\) denotes a cyclic
spatial shift by \(\boldsymbol{\delta}\) pixels. A subsequent
{spectral attenuation} step scales each channel with
asymmetric gains to mimic underwater absorption, where red is
preferentially suppressed and blue slightly boosted:
\begin{equation}
  \tilde{\mathbf{I}}_d = \mathrm{clamp}\!\bigl(
  \operatorname{diag}(\alpha_R,\alpha_G,\alpha_B)\,\mathbf{I}'_{d},\;0,\;1\bigr).
\end{equation}

Both transformations are applied jointly with probability \(p{=}0.8\)
and disabled at inference. 
In addition, a lightweight internal augmentation generates weakly and
strongly perturbed views of the prediction for the self-supervised
consistency loss defined in Eq.~\eqref{eq:ssl_loss}.

\subsection{Gray-Edge Prior Module}\label{sec:grayedge}
Underwater images often suffer from global colour casts due to
wavelength-dependent absorption. We therefore derive a Gray-Edge
prior~\cite{Gray_Edge} as a conditioning signal. Unlike Gray-World, which
equalises channel means, GEPM operates on image derivatives and assumes
achromatic average edge magnitudes across channels.

Using Scharr gradients, we compute the channel-wise gains as follow:
\begin{equation}\label{eq:grayedge}
    g_c =
    \frac{\bar{e}}
    {\bigl(\mathbb{E}[\lVert \nabla_c \mathbf{I} \rVert^p]\bigr)^{1/p}
    + \epsilon},
    \quad c \in \{R,G,B\},
\end{equation}
where \(\bar{e}\) denotes the mean statistic across channels, \(p\) is the
Minkowski exponent, and \(\epsilon\) ensures numerical stability. The gains are
clamped to \([0.5,2.5]\), followed by log-domain mean correction and
re-normalisation to obtain \(\mathbf{I}_{\mathrm{wb}}\). A learnable
channel-wise parameter then blends it with the original input:
\begin{equation}\label{eq:wb_blend}
    \mathbf{\tilde{I}}_d =
    \sigma(\boldsymbol{\alpha}) \odot \mathbf{I}_{\mathrm{wb}}
    + \bigl(1-\sigma(\boldsymbol{\alpha})\bigr)
    \odot \mathbf{I}_d ,
\end{equation}
where
\(\boldsymbol{\alpha}\in\mathbb{R}^{1\times3\times1\times1}\)
controls the correction strength of each channel.

\subsection{Guided Dynamic Feature Modulation Block}\label{sec:GDFMB}

The GDFMB accepts two inputs: the guide image \(\mathbf{\tilde{I}}_{{d}} \in \mathbb{R}^{3 
\times H \times W}\) and the intermediate feature map 
\(\mathbf{F}_{in} \in \mathbb{R}^{C \times H' \times W'}\). Its output is a modulated 
feature map of the same shape as \(\mathbf{F}_{out}\). The GDFMB comprises two core components: the Guided Degradation-Adaptive Block (GDAB) and the Gradient Fusion Block (GFB), each described in detail below.

\subsubsection{Guided Degradation-Adaptive Block.}
The guide image and feature map are projected by $3\times3$ convolutions into a shared hidden space of dimension $D=2C$, yielding $\mathbf{X}\in\mathbb{R}^{D\times H\times W}$ and $\mathbf{Y}\in\mathbb{R}^{D\times H'\times W'}$. GDAB then performs deformable cross-attention with $\mathbf{Y}$ as the query and $\mathbf{X}$ as the key-value.
The query $\mathbf{Q}=W_q\mathbf{Y}$ is divided into $G$ groups, each predicting a 2-D offset field:
\begin{equation}\label{eq:offset}
    \Delta\mathbf{p}_g =
    \tanh\!\bigl(\mathrm{OffsetNet}_g(\mathbf{Q}_g)\bigr)
    \odot \boldsymbol{\rho}\cdot r,
\end{equation}
where $\boldsymbol{\rho}$ normalises the offsets according to resolution and $r$ controls their range. The offset network is zero-initialised. Guide features are bilinearly sampled at
$\mathbf{p}=\mathbf{p}_{\mathrm{ref}}+\Delta\mathbf{p}$ and used as the key-value pair in multi-head cross-attention. This allows the module to retrieve spatially varying degradation cues relevant to each feature location.

Global average pooling then produces a feature embedding
$\mathbf{e}_f\in\mathbb{R}^{D}$ and a degradation embedding
$\mathbf{e}_d\in\mathbb{R}^{D}$. After projection to
$\tilde{\mathbf{e}}_f\in\mathbb{R}^{D/2}$, each element is expanded over $K$ Gaussian basis functions:
\begin{equation}\label{eq:gbe}
    \phi_k(x)=
    \exp\!\left(-\frac{(x-\mu_k)^2}{\delta^2}\right),
    \quad k=1,\ldots,K,
\end{equation}
where $\mu_k$ are uniformly spaced centres and
$\delta=(v_{\max}-v_{\min})/(K-1)$. The resulting basis representation is dynamically modulated using weights generated jointly from $\mathbf{e}_d$ and $\tilde{\mathbf{e}}_f$, producing
$\mathbf{m}_{\mathrm{GBE}}\in\mathbb{R}^{D}$.

A selective fusion gate blends $\mathbf{m}_{\mathrm{GBE}}$ with the original feature embedding:
\begin{equation}\label{eq:sel_gate}
    \omega_{\mathrm{emb}},\omega_{\mathrm{GBE}}
    =
    \mathrm{softmax}\!\bigl(
    [\mathbf{h}_{\mathrm{emb}},\mathbf{h}_{\mathrm{GBE}}]
    \bigr),
\end{equation}
\begin{equation}\label{eq:sel_fusion}
    \mathbf{m}
    =
    \omega_{\mathrm{emb}}\odot\mathbf{e}_f
    +
    \omega_{\mathrm{GBE}}\odot\mathbf{m}_{\mathrm{GBE}},
\end{equation}
where the two scores are predicted by parallel linear heads. Finally, $\mathbf{m}$ modulates the spatial GDAB features and is injected into the main branch through channel-wise cross-attention.

\subsubsection{Gradient Fusion Block.}
To preserve high-frequency structures attenuated by underwater degradation, we
append a GFB after each GDFMB. Given an input feature \(\mathbf{F}\), a
channel-wise Sobel operator computes the gradient magnitude
\[
\mathbf{G}=\sqrt{\mathbf{G}_x^2+\mathbf{G}_y^2+\epsilon},
\]
which is transformed by a depthwise-separable convolution and sigmoid activation
to obtain a spatial mask \(\mathbf{M}_{\mathrm{grad}}\in[0,1]\). After an initial
ConvsBlock, the feature is modulated as
\begin{equation}\label{eq:sgfb}
    \mathbf{F}_{out}
    =
    \sigma(\alpha)\,\mathbf{M}_{\mathrm{grad}}\odot\mathbf{F}_1
    +
    \bigl(1-\sigma(\alpha)\bigr)\mathbf{F}_1,
\end{equation}
where \(\mathbf{F}_1\) denotes the refined feature and \(\alpha\) is a learnable
scalar controlling the gradient contribution. A subsequent BasicBlock further
refines the output, enabling content-adaptive preservation of edges and textures.

\subsection{Task-Oriented Latent Control}\label{sec:latent}

\subsubsection{Mapper.}
At the encoder bottleneck, the deepest feature
\(\mathbf{F}_{l}\in\mathbb{R}^{4C\times\frac{H}{4}\times\frac{W}{4}}\)
contains both scene and degradation information. A Mapper
\(\mathcal{M}\), implemented by two cascaded ConvsBlocks, produces a
task-oriented control code:
\begin{equation}\label{eq:latent_mapper}
    \mathbf{F}_{z}=\mathcal{M}(\mathbf{F}_{l}), \quad
    \mathbf{F}_{res}=\mathbf{F}_{l}-\mathbf{F}_{z}.
\end{equation}
Here, \(\mathbf{F}_{z}\) serves as the bottleneck feature and controls the
skip-feature gates, while \(\mathbf{F}_{res}\) is its complementary residual.
We do not interpret them as a physical clean/degradation decomposition;
instead, \(\mathbf{F}_{z}\) is defined by its role in producing
restoration-oriented channel modulation.

\subsubsection{Task-Oriented Latent Regularisation.}
Reconstruction supervision alone does not explicitly structure the bottleneck
representation used for skip modulation. Inspired by BaryIR~\cite{BaryIR}, we
therefore introduce a composite regularisation
$\mathcal{L}_{\mathrm{center}}$ to learn a discriminative control code
$\mathbf{F}_{z}$. These objectives act as optimisation biases rather than
enforcing a clean-manifold projection.

{1) Reference-Latent Guidance.}
We align $\mathbf{F}_{z}$ with the representation
$\mathbf{F}_{z}^{gt}$ encoded from the ground-truth image:
\begin{equation}\label{eq:align_loss}
    \mathcal{L}_{\mathrm{align}}
    =
    \left\|
    \mathbf{F}_{z}-\mathbf{F}_{z}^{gt}
    \right\|_2^2.
\end{equation}
This term anchors the control code to restoration-relevant information.

{2) Orthogonality Regularisation.}
We reduce redundancy between the control code and its residual
$\mathbf{F}_{res}$ through
\begin{equation}\label{eq:orth_loss}
    \mathcal{L}_{\mathrm{orth}}
    =
    \frac{1}{B}\sum_{i=1}^{B}
    \left|
    \left\langle
    \hat{\mathbf{F}}_{z}^{(i)},
    \hat{\mathbf{F}}_{res}^{(i)}
    \right\rangle
    \right|,
\end{equation}
where $\hat{\mathbf{F}}=\mathbf{F}/\|\mathbf{F}\|_2$ denotes the normalised
flattened feature. This term encourages complementary representations without
assigning them explicit clean or degradation semantics.

{3) Contrastive Regularisation.}
Given negative latents
$\mathcal{N}=\{\mathbf{F}_{z}^{neg(n)}\}_{n=1}^{N}$, we use
\begin{equation}\label{eq:contrast_loss}
    \mathcal{L}_{\mathrm{contrast}}
    =
    \frac{1}{|\mathcal{N}|}
    \sum_{n=1}^{|\mathcal{N}|}
    \max\!\left(
    0,\,
    d\!\left(\hat{\mathbf{F}}_{z},
             \hat{\mathbf{F}}_{z}^{gt}\right)
    -
    d\!\left(\hat{\mathbf{F}}_{z},
             \hat{\mathbf{F}}_{z}^{neg(n)}\right)
    +m
    \right),
\end{equation}
which separates the control code from degraded negatives and discourages
representation collapse.

The overall regularisation is
\begin{equation}\label{eq:latent_loss}
    \mathcal{L}_{\mathrm{center}}
    =
    \lambda_a\mathcal{L}_{\mathrm{align}}
    +
    \lambda_o\mathcal{L}_{\mathrm{orth}}
    +
    \lambda_c\mathcal{L}_{\mathrm{contrast}},
\end{equation}
where $(\lambda_a,\lambda_o,\lambda_c)=(0.1,0.05,0.05)$.
Together, the three terms structure $\mathbf{F}_{z}$ as a task-oriented
control representation for skip modulation.

\subsubsection{Attention Gate.}
To regulate skip-feature reuse, each skip connection is equipped with an
Attention Gate. Channel-wise weights are generated from the control code
\(\mathbf{F}_{z}\) using adaptive average pooling and a two-layer MLP:
\begin{equation}\label{eq:att_gate}
    \mathbf{F}_{s}
    =
    \mathbf{F}_{out}^{enc}
    \odot
    \sigma\!\bigl(
    \mathrm{MLP}(\mathrm{GAP}(\mathbf{F}_{z}))
    \bigr).
\end{equation}
The resulting latent-conditioned modulation rescales skip channels before they
are fused with the upsampled decoder features. The weights act as reconstruction controls rather than degradation estimates.

\begin{table*}[!htb]
  \centering

  \resizebox{\textwidth}{!}
  {
  \begin{tabular}{l|ccc|ccc|cc}
    \toprule
    \multirow{2}{*}{\bf Methods} 
    & \multicolumn{3}{c|}{\bf U90} 
    & \multicolumn{3}{c|}{\bf LSUI} 
    & \multirow{2}{*}{Params (M)$\downarrow$} & \multirow{2}{*}{FLOPs (G)$\downarrow$}
    \\
    
    \cmidrule(lr){2-4} \cmidrule(lr){5-7}
    & PSNR$\uparrow$ & SSIM$\uparrow$  & LPIPS$\downarrow$ 
    & PSNR$\uparrow$ & SSIM$\uparrow$  & LPIPS$\downarrow$ 
    & &
    \\
    
    \midrule
    WaterNet (TIP'19) \cite{UIEB}              
    & 20.87 & 0.911 & 0.157
    & 23.07 & 0.853 & 0.221
    & 24.81 & 193.70
    \\

    Ucolor (TIP'21) \cite{Ucolor}              
    & 20.81 & 0.904 & 0.160 
    & 23.28 & 0.890 & 0.211 
    & 157.40 & 34.68
    \\

    MLLE (TIP'22) \cite{MLLE}              
    & 18.68 & 0.855 & 0.286 
    & 22.23 & 0.835 & 0.254 
    & - & -
    \\
    
    Ushape (TIP'23) \cite{Ushape}              
    & 20.97 & 0.864 & 0.203 
    & 24.46 & 0.901 & 0.190 
    & 65.60 & 66.20
    \\
    
    NU2Net (AAAI'23) \cite{NU2Net}              
    & 22.91 & 0.922 & 0.174 
    & 25.32 & 0.910 & 0.162 
    & 3.10 & 10.40
    \\

    Semi-UIR (CVPR'23) \cite{Semi-UIR}              
    & 24.19 & 0.923 & 0.151 
    & 27.08 & 0.907 & 0.134 
    & 12.78 & 36.46
    \\

    GUPDM (MM'23) \cite{GUPDM} 
    & 24.33 & \underline{0.928} & 0.133 
    & 27.77 & 0.917 & 0.138 
    & 5.60  & 191.61
    \\
    
    HCLR-Net (IJCV'24)  \cite{zhou_IJCV_HCLR}             
    & 23.72 & 0.917 & 0.158 
    & 26.98 & 0.904 & 0.130 
    & 4.87 & 401.97
    \\

UniUIR (TIP'25) \cite{UniUIR}
    & \underline{25.11} & \textbf{0.933} & 0.112
    & 28.42 & \textbf{0.926} & 0.123
    & 29.19 & 78.07
    \\

    AdaIR (ICLR'25)  \cite{AdaIR}
    & 24.16 & 0.926 & 0.131
    & 27.55 & \underline{0.920} & 0.139 
    & 28.77 & 147.50
    \\

    MoCE-IR (CVPR'25) \cite{MoCE-IR}              
    & 24.93 & 0.926 & 0.123 
    & 28.19 & 0.910 & 0.124 
    & 25.35 & 103.03
    \\

    WWE-UIE (WACV'26) \cite{WWE-UIE}                
    & 24.17 & 0.926 & \underline{0.086}
    & \underline{28.62} & 0.913 & \underline{0.085}
    & 0.73  & 6.25
    \\
    
    \rowcolor[HTML]{EDEDED}
    \textbf{PROTEUS} (Ours)               
    & \textbf{25.50} & \textbf{0.933}  & \textbf{0.081} 
    & \textbf{28.99} & 0.913  & \textbf{0.083}   
    & 2.61  & 18.52
    \\

    
    \bottomrule
  \end{tabular}
  }
  \caption{Quantitative results on {referenced-based} datasets: U90 and LSUI-400.}
  \label{tab:uie_benchmark}
      \vspace{-1em}
\end{table*}

\subsection{Loss Functions}\label{sec:loss_funcs}

The training objective contains five standard restoration losses adopted from
WWE-UIE~\cite{WWE-UIE}: the Charbonnier loss
$\mathcal{L}_{\ell_1}$, HVI colour loss
$\mathcal{L}_{\mathrm{HVI}}$~\cite{CIDNet}, SSIM loss
$\mathcal{L}_{\mathrm{SSIM}}$, VGG perceptual loss
$\mathcal{L}_{\mathrm{VGG}}$, and edge loss
$\mathcal{L}_{\mathrm{edge}}$. We additionally introduce latent regularisation
and self-supervised consistency.

\subsubsection{Latent Control Regularisation.}
The loss $\mathcal{L}_{\mathrm{center}}$ in
Eq.~\eqref{eq:latent_loss} structures the control code through reference
alignment, orthogonality, and contrastive separation. Negative latents are
obtained from precomputed outputs of BaryIR~\cite{BaryIR},
FUnIE~\cite{FUnIE}, and USUIR~\cite{USUIR}.

\subsubsection{Self-Supervised Consistency.}
We impose consistency between weakly and strongly augmented views of the
network prediction:
\begin{equation}\label{eq:ssl_loss}
    \mathcal{L}_{\mathrm{ssl}}
    =
    \beta(t)
    \left\|
    \mathbf{I}_{r}^{c}-\mathbf{I}_{r}^{b}
    \right\|_2^2,
\end{equation}
where
$\beta(t)=\frac{\beta_0}{2}\bigl(\cos(\pi t/T)+1\bigr)$
decays from $\beta_0=0.1$ to zero during training.

\subsubsection{Total Loss.}
The overall objective is
\begin{equation}\label{eq:total_loss}
\begin{aligned}
    \mathcal{L} =\;&
    w_{\ell_1}\mathcal{L}_{\ell_1}
    +w_{\mathrm{HVI}}\mathcal{L}_{\mathrm{HVI}}
    +w_{\mathrm{SSIM}}\mathcal{L}_{\mathrm{SSIM}}
    +w_{\mathrm{VGG}}\mathcal{L}_{\mathrm{VGG}} \\
    &+w_{\mathrm{edge}}\mathcal{L}_{\mathrm{edge}}
    +w_{\mathrm{center}}\mathcal{L}_{\mathrm{center}}
    +w_{\mathrm{ssl}}\mathcal{L}_{\mathrm{ssl}},
\end{aligned}
\end{equation}
where $w_{\ell_1} = 1.0$, $w_{{HVI}} = 0.5$, $w_{{SSIM}} = 0.1$, $w_{{VGG}} = 0.1$, $w_{{edge}} = 0.1$, $w_{{center}} = 1.0$, and $w_{{ssl}} = 1.0$.

\section{Experiments}

\subsection{Experimental Setup}

\subsubsection{Datasets.} We evaluate on five paired benchmarks including U90~\cite{UIEB}, LSUI~\cite{Ushape}, UFO~\cite{UFO}, EUVP-Scene, and EUVP-Dark~\cite{EUVP}, alongside four non-reference benchmarks: Challenge-60~\cite{UIEB}, U45~\cite{U45}, UCCS~\cite{UCCS}, and EUVP-330~\cite{EUVP}. Dataset splits are provided in Supplementary Materials.

\subsubsection{Metrics.} We use PSNR, SSIM~\cite{SSIM}, and LPIPS~\cite{LPIPS} for paired data, and UCIQE~\cite{UCIQE}, UIQM~\cite{UIQM}, and URanker~\cite{NU2Net} for non-reference evaluation.

\subsubsection{Implementation Details.} 
PROTEUS is implemented in PyTorch and trained end-to-end with AdamW
($\beta_1=0.9$, $\beta_2=0.999$, weight decay $10^{-6}$). The learning rate is
linearly warmed up from zero to $1\times10^{-4}$ over the first 10 epochs and
kept constant thereafter. We train for 1,000 epochs on UIEB and 300 epochs on
the other datasets, using randomly cropped $256\times256$ patches, a per-GPU
batch size of 2, and four NVIDIA RTX 4090 GPUs. Gradient norms are clipped at
1.0. Photometric augmentation simulates wavelength-dependent attenuation via
random spectral gains and sub-pixel channel shifts, while geometric
augmentation imposes consistency under stochastic transformations.

\begin{table*}[!tb]
  \centering
  \resizebox{\textwidth}{!}
  {
  \begin{tabular}{l|cc|cc|cc|cc}
    \toprule
    \multirow{2}{*}{\bf Methods} &
    \multicolumn{2}{c|}{\bf UFO-120} &
    \multicolumn{2}{c|}{\bf EUVP-S} &
    \multicolumn{2}{c|}{\bf EUVP-D}
    & \multirow{2}{*}{Params (M)$\downarrow$} & \multirow{2}{*}{FLOPs (G)$\downarrow$}
    \\
    
    \cmidrule(lr){2-3}\cmidrule(lr){4-5}\cmidrule(lr){6-7}
    & PSNR$\uparrow$ & SSIM$\uparrow$
    & PSNR$\uparrow$ & SSIM$\uparrow$
    & PSNR$\uparrow$ & SSIM$\uparrow$
    & &
    \\
    
    \midrule
    NU2Net (AAAI'23) \cite{NU2Net}             
    & 16.23 & 0.674 
    & 24.28 & 0.786 
    & 21.84 & 0.891
    & 3.15  & 10.49
    \\
    
    Ushape (TIP'23) \cite{Ushape}              
    & 26.65 & 0.820 
    & 24.93 & 0.746 
    & 21.69 & 0.876 
    & 65.60 & 66.20
    \\
    
    SMDR-IS (AAAI'24) \cite{SMDR-IS}            
    & \underline{28.26}   & 0.860 
    & 26.50   & 0.809 
    & 22.39   & \textbf{0.909} 
    & 12.25   & 48.32
    \\
    
    CDF-UIE (TGRS'25) \cite{CDF-UIE}            
    & 27.76   & 0.855 
    & 26.37   & 0.808 
    & 22.05   & 0.896 
    & 15.90   & 57.46
    \\
    
    Phaseformer (WACV'25) \cite{Phaseformer}        
    & 27.44 & 0.807 
    & 25.90 & 0.763
    & 22.12 & 0.895 
    & 1.78  & 13.04
    \\
    
    MoCE-IR (CVPR'25) \cite{MoCE-IR}
    & 28.09 & 0.876
    & 27.15 & 0.832
    & 22.28 & 0.905
    & 25.35 & 103.03
    \\

    UniUIR (TIP'25) \cite{UniUIR}  
    & 27.97 & \underline{0.883}
    & \underline{27.56} & \underline{0.849}
    & 22.34 & 0.904 
    & 29.19 & 78.07
    \\

    WWE-UIE (WACV'26) \cite{WWE-UIE}                
    & \textbf{28.28} & 0.867 
    & 26.59 & 0.824 
    & \underline{22.53} & \underline{0.908} 
    & 0.73  & 6.25
    \\

    \rowcolor[HTML]{EDEDED}
    \textbf{PROTEUS} (Ours)
    & \underline{28.26}   & \textbf{0.894}
    & \textbf{28.39}   & \textbf{0.893} 
    & \textbf{22.68}   & \underline{0.908}
    & 2.61    & 18.52
    \\
    
    \bottomrule
  \end{tabular}
  }
  \caption{Quantitative results on {referenced-based} datasets: UFO-120, EUVP-Scene, and EUVP-Dark.}
  \label{tab:ufo}
      \vspace{-0.5em}
\end{table*}

\subsection{Main Results}

\subsubsection{Results on Referenced Datasets.} 
\begin{figure*}[!t]
\centering
\includegraphics[page=1,width=0.9\textwidth]{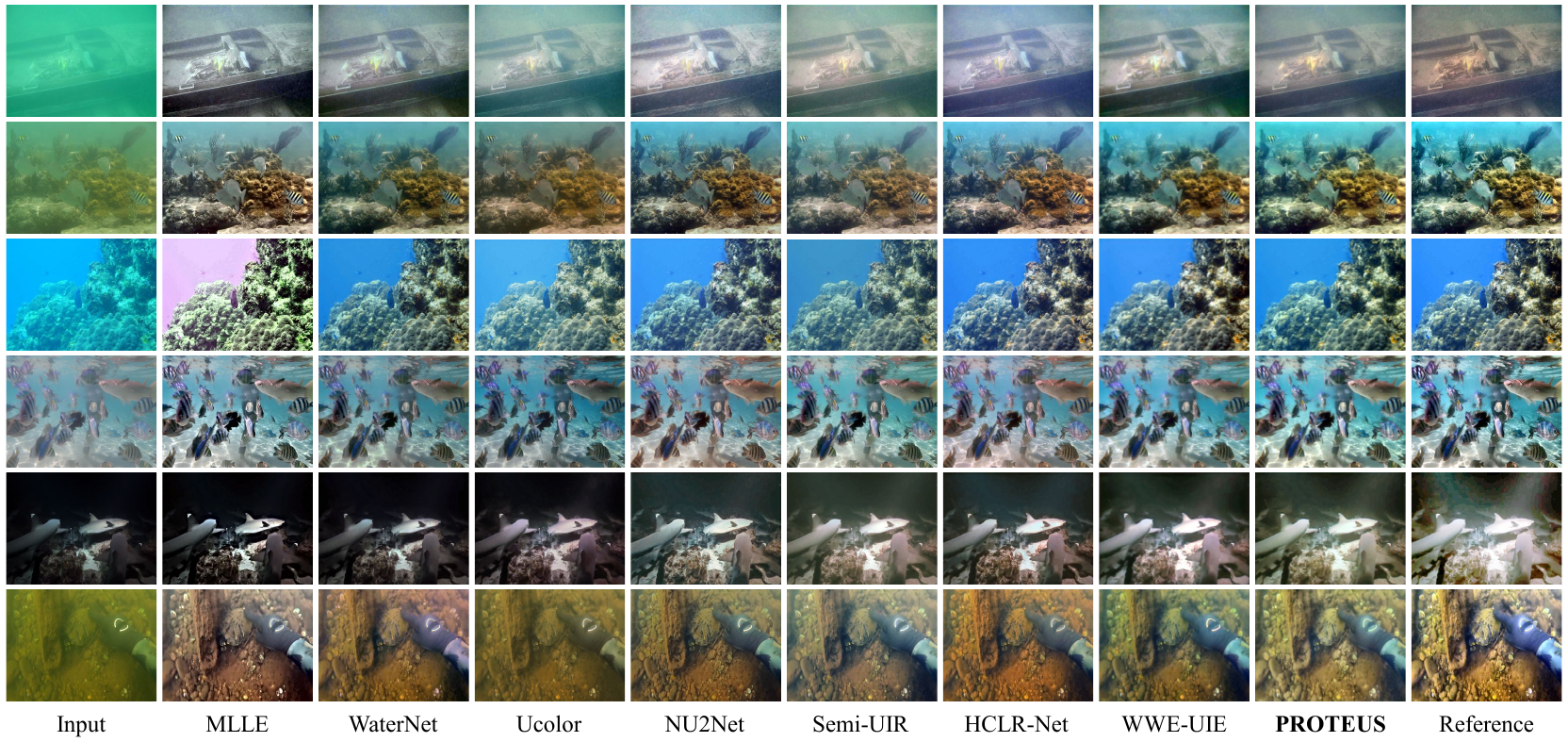}
\caption{Full visual comparison on U90~\cite{UIEB} dataset.}
\label{fig:UIEB_full}
\vspace{-1em}
\end{figure*}

Tables~\ref{tab:uie_benchmark} and~\ref{tab:ufo} report quantitative
comparisons on five referenced benchmarks. On U90, PROTEUS achieves the best
PSNR and LPIPS together with joint-best SSIM, while on LSUI-400 it obtains the
best PSNR and LPIPS. On UFO-120, it achieves the best SSIM and joint-second
PSNR, only 0.02\,dB below WWE-UIE. It ranks first in both metrics on
EUVP-Scene; on EUVP-Dark, it attains the best PSNR and joint-second SSIM,
only 0.001 below the top-ranked method.
PROTEUS also retains the best LPIPS, improving over the second-ranked WWE-UIE by approximately 6\% on U90 and 2\% on LSUI. 
The representative outcome in Fig.~1(c) shows that PROTEUS reduces residual
colour casts and improves local contrast compared with a representative
competing method.
Full visual comparisons are provided in Fig.~\ref{fig:UIEB_full} and
Supplementary Materials.

\begin{table*}[!htb]
\centering
\renewcommand{\arraystretch}{1}
\setlength\tabcolsep{1pt}

\resizebox{\textwidth}{!}{
\begin{tabular}{l| ccc |ccc| ccc | ccc}
\toprule
\multirow{2}{*}{Method}
  & \multicolumn{3}{c|}{U45}
  & \multicolumn{3}{c|}{Challenge-60}
  & \multicolumn{3}{c|}{UCCS}
  & \multicolumn{3}{c}{EUVP-330}
  \\
  
\cmidrule(lr){2-4}\cmidrule(lr){5-7}\cmidrule(lr){8-10}\cmidrule(lr){11-13}
& UCIQE$\uparrow$ & UIQM$\uparrow$ & URanker$\uparrow$
& UCIQE$\uparrow$ & UIQM$\uparrow$ & URanker$\uparrow$
& UCIQE$\uparrow$ & UIQM$\uparrow$ & URanker$\uparrow$
& UCIQE$\uparrow$ & UIQM$\uparrow$ & URanker$\uparrow$ 
\\

\midrule
WaterNet (TIP'19) \cite{UIEB}
& 0.572 & 3.195 & 1.312
& 0.566 & 2.653 & 1.223
& 0.545 & 3.058 & 1.281
& 0.524 & 3.019 & 1.685 \\

Ucolor (TIP'21) \cite{Ucolor}
& 0.564 & 3.151 & 1.485
& 0.532 & 2.746 & 1.213
& 0.550 & 3.019 & 1.064
& 0.561 & 3.114 & 1.546 \\

MLLE (TIP'22) \cite{MLLE}
& 0.593 & 2.599 & 1.349
& 0.581 & 2.310 & 1.257
& 0.544 & 2.985 & 1.145
& 0.568 & 3.026 & 1.629 \\

Ushape (TIP'23) \cite{Ushape}
& 0.553 & 3.048 & 1.737
& 0.534 & 2.783 & 1.363
& 0.567 & 3.012 & 1.374
& 0.557 & 3.013 & 1.679 \\

NU2Net (AAAI'23) \cite{NU2Net}
& 0.595 & 3.206 & 1.804
& 0.564 & 2.907 & 1.557
& \textbf{0.601} & 2.994 & \underline{1.624}
& 0.571 & 3.127 & 1.801 \\

Semi-UIR (CVPR'23) \cite{Semi-UIR}
& 0.601 & \underline{3.323} & 1.831
& 0.573 & 2.925 & 1.685
& 0.552 & 3.039 & 1.321
& 0.526 & 2.994 & 1.864 \\

GUPDM (MM'23) \cite{GUPDM}
& 0.566 & 3.057 & 1.872
& 0.548 & 2.687 & 1.751
& \underline{0.586} & 3.157 & 1.583
& 0.563 & 3.014 & 1.733 \\

HCLR-Net (IJCV'24) \cite{zhou_IJCV_HCLR}
& 0.585 & 3.013 & 1.927
& 0.569 & 2.737 & 1.615
& 0.541 & 3.009 & \textbf{1.784}
& 0.559 & 3.102 & \textbf{2.052} \\

UniUIR (TIP'25) \cite{UniUIR}
& \textbf{0.609} & 3.278 & \underline{2.028}
& \underline{0.593} & 3.218 & \textbf{1.825}
& 0.576 & 3.215 & 1.603
& 0.581 & \underline{3.198} & 1.927 \\

AdaIR (ICLR'25) \cite{AdaIR}
& 0.582 & 3.195 & 1.852
& {0.587} & 2.942 & 1.706
& 0.559 & 3.073 & 1.564
& 0.560 & 3.105 & 1.823 \\

MoCE-IR (CVPR'25) \cite{MoCE-IR}
& 0.591 & 3.204 & {1.933}
& \textbf{0.594} & 3.014 & \underline{1.779}
& 0.585 & 3.035 & 1.421
& 0.582 & 3.068 & {1.807} \\

WWE-UIE (WACV'26) \cite{WWE-UIE}
& \underline{0.605} & \textbf{3.326} & 2.000
& 0.570 & \underline{3.368} & 1.497
& 0.537 & \textbf{3.264} & 1.392
& \underline{0.585} & 3.127 & 1.915 \\

\rowcolor[HTML]{EDEDED}
\textbf{PROTEUS} (Ours)
& 0.604 & 3.270 & \textbf{2.061}
& 0.575 & \textbf{3.398} & 1.593
& 0.558 & \underline{3.248} & 1.584
& \textbf{0.593} & \textbf{3.213} & \underline{1.959} \\

\bottomrule
\end{tabular}
}
\caption{Quantitative results on {non-referenced-based} datasets: U45, Challenge-60, UCCS, and EUVP-330.}
\label{tab:non_ref}
    \vspace{-0.5em}
\end{table*}

\noindent\textbf{Results on Non-Referenced Datasets.} 
Table~\ref{tab:non_ref} evaluates four non-reference benchmarks. PROTEUS ranks first in four and second in two of the twelve metric--dataset combinations, including the best URanker on U45, best UIQM on Challenge-60, and best UCIQE and UIQM on EUVP-330. These results, together with the qualitative comparisons in Supplementary Materials, demonstrate robust generalisation to real-world underwater images.

\begin{table}[!htb]
\centering

\resizebox{\linewidth}{!}{
\begin{tabular}{l|ccc}

\toprule
\bf Configuration  & PSNR$\uparrow$ & SSIM$\uparrow$ & LPIPS$\downarrow$
\\

\midrule
w/o Photometric \& Geometric Aug.
& 25.31 & 0.928 & 0.096
\\

w/o GEPM
& 25.30 & 0.929 & 0.092
\\

w/o Attention Gate
& {25.08} & {0.921} & {0.107}
\\

w/o Mapper 
& {25.19} & {0.929} & {0.094}
\\

w/o GDAB
& 25.27 & 0.927 & 0.096
\\

w/o deformable offset 
& {25.10} & {0.924} & {0.104}
\\

w/o Gaussian Basis Expansion 
& {25.15} & {0.922} & {0.103}
\\

w/o GFB
& \underline{25.36} & \underline{0.930} & 0.093
\\

CBAM instead of Attention Gate
& 25.34 & 0.929 & \underline{0.091}
\\

Restormer Block instead of GDFMB
& 25.21 & 0.926 & 0.095
\\

\textbf{Full Model}
& \textbf{25.50} & \textbf{0.933} & \textbf{0.081}
\\

\bottomrule
\end{tabular}
}
\caption{Ablation study on module components.}
\label{tab:ablation_component}
\end{table}

\subsection{Ablation Study}

We performed a series of ablation studies to validate the contribution of individual components in the proposed method, with all experiments conducted on the UIEB dataset.

\noindent\textbf{Ablation Study on Module Components.}
Table~\ref{tab:ablation_component} shows that the full model performs best on all three metrics. Removing the Attention Gate, Mapper, deformable offset, or Gaussian Basis Expansion lowers PSNR from 25.50\,dB to 25.08, 25.19, 25.10, and 25.15\,dB, respectively, with corresponding SSIM and LPIPS degradation. All remaining component removals or replacements also underperform the full model, consistently supporting each design choice. Detailed loss ablations are provided in Supplementary Materials.

\begin{figure}[!t]
\centering
\includegraphics[width=1\columnwidth]{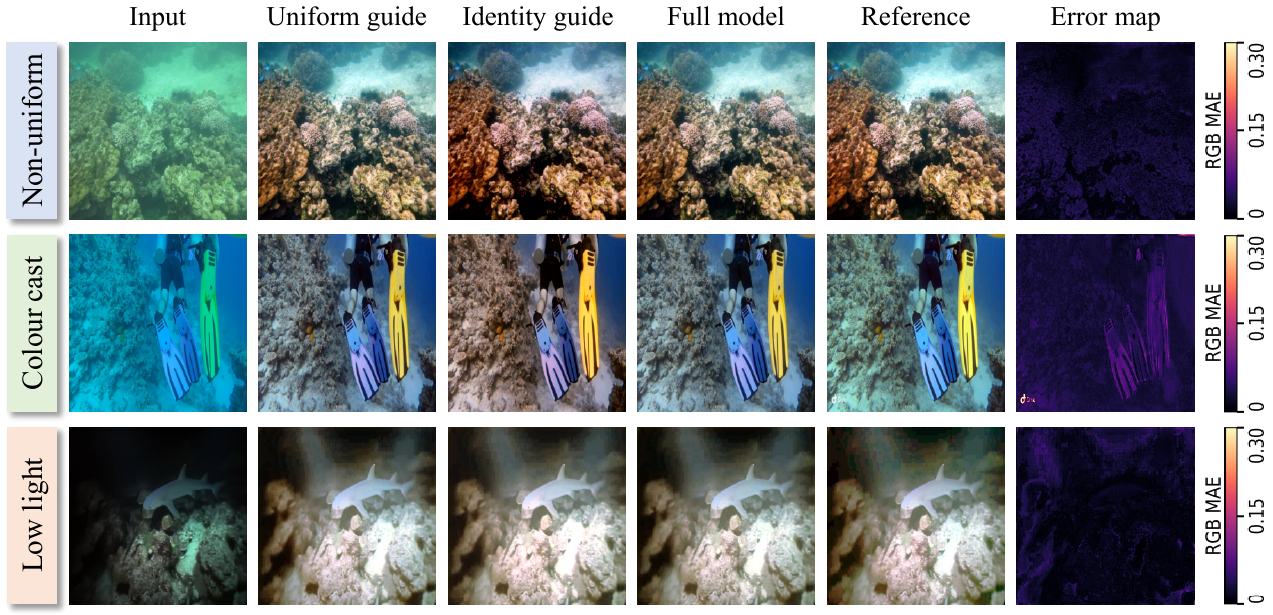}
\caption{Visual analysis on U90 dataset. ``Uniform guide”
replaces the spatially varying guide supplied to each GDFMB
with its per-channel spatial mean, while ``Identity gate” replaces both latent-conditioned skip gates with all-one vectors.
These interventions isolate sensitivity to the two controls.}
\label{fig:visual_ablation}
\vspace{-1em}
\end{figure}

\begin{figure}[!t]
\centering
\includegraphics[width=0.90\columnwidth]{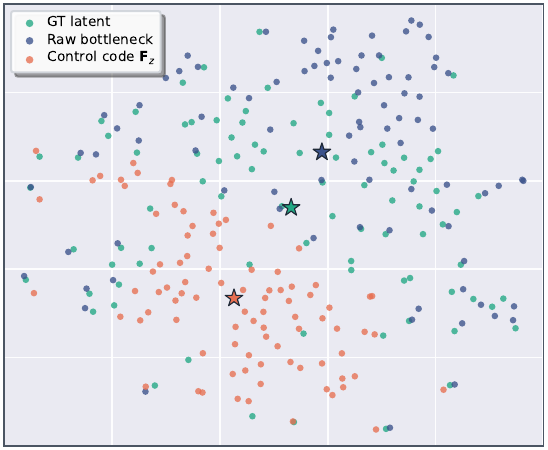}
\captionof{figure}{{Joint t-SNE of raw bottlenecks, control codes $\mathbf{F}_z$, and ground-truth latents on U90. Colours identify the representations; stars mark their 2D centroids. This descriptive projection supports task-oriented remapping, not clean/degradation separation; proximity claims use the original-space diagnostics.}}
\label{fig:latent_tsne}
\vspace{-1em}
\end{figure}

\subsubsection{Visual Analysis of Dual Control.}
Figure~\ref{fig:visual_ablation} further validates the two complementary
controls illustrated in Fig.~\ref{fig:motivation}(b).
Replacing spatially varying guidance with a uniform guide produces visible colour or illumination shifts, while replacing the learned skip gates with identity weights causes broader colour and contrast deviations around foreground structures. Combining both controls yields outputs closer to the references in these examples. Because the gates apply global channel-wise weights, these results demonstrate complementary reconstruction sensitivity rather than spatial localisation of degradation.

\begin{table}[t]
\centering

\begin{tabular}{lccc}
\toprule
\textbf{Gate setting} &
{PSNR$\uparrow$} &
{SSIM$\uparrow$} &
{LPIPS$\downarrow$} \\
\midrule
Predicted gate            & 25.50 & 0.933 & 0.081 \\
All-one gate              & 22.15 & 0.887 & 0.121 \\
Fixed mean gate           & 25.23 & 0.931 & 0.089 \\
Channel-shuffled gate     & 24.43 & 0.925 & 0.094 \\
Cross-image gate          & 25.08 & 0.928 & 0.090 \\
\bottomrule
\end{tabular}
\caption{Test-time skip-gate interventions on the 90 U90 pairs. All network
weights remain fixed. The fixed gate is estimated from the disjoint UIEB
training split, while channel-shuffled and cross-image results are averaged
over five deterministic repeats per image.}
\label{tab:gate_intervention}
\vspace{-1em}
\end{table}

\subsubsection{Representation and Gate Diagnostics.}
Across 90 U90 pairs, $\mathbf{F}_{z}$ and $\mathbf{F}_{res}$ are nearly
orthogonal, with a mean absolute cosine similarity of 0.0077, while the gates
attenuate skip features by 23.4\% on average. Nevertheless, $\mathbf{F}_{z}$
is farther from the ground-truth latent than the unfiltered bottleneck
(MSE: 0.0164 vs.\ 0.0145; cosine: 0.7162 vs.\ 0.8067), supporting its role as
a task-oriented control code rather than a clean-space projection. The three-seed analysis confirms this conclusion; visualisations are provided
in the Supplementary Material.

Table~\ref{tab:gate_intervention} further shows that replacing the predicted
gates with all-one vectors reduces PSNR by 3.35\,dB, while shuffling their
channel assignments causes a 1.07\,dB drop. Fixed-mean and cross-image gates
also reduce PSNR by 0.27 and 0.42\,dB, respectively. These results indicate
that the controller captures a stable channel-modulation structure while
retaining meaningful input-specific adjustments, rather than directly
detecting degradation-affected channels.

\subsubsection{Representation Geometry.}
Figure~\ref{fig:latent_tsne} shows that $\mathbf{F}_z$ occupies a
distribution distinct from the raw and ground-truth latents, indicating
substantial task-oriented remapping. This pattern should not be
interpreted as clean-latent convergence, consistent with the
original-space MSE and cosine diagnostics. We therefore use t-SNE only
as a qualitative visualization of representation restructuring.

\section{Conclusion}

This work presents PROTEUS, which exploits informative degradation cues for
adaptive feature processing while regulating degradation-entangled
skip-feature reuse through task-oriented latent control. GDFMB uses spatially
varying degradation cues to guide feature adaptation, while the latent
controller regulates encoder-feature reuse during decoding.
Experiments on five paired and four non-reference benchmarks demonstrate competitive restoration performance with 2.61\,M parameters and 18.52\,G FLOPs.

\bibliography{MM26_UIE}

@String(IJCV  = {IJCV})

@String(CVPR  = {IEEE Conf. Comput. Vis. Pattern Recog.})

@String(ICCV  = {Int. Conf. Comput. Vis.})

@String(ECCV  = {Eur. Conf. Comput. Vis.})

@String(NeurIPS = {Adv. Neural Inform. Process. Syst.})

@String(ICLR  = {Int. Conf. Learn. Represent.})

@String(BMVC  = {Brit. Mach. Vis. Conf.})

@String(AAAI  = {AAAI})

@String(TIP   = {IEEE TIP})

@String(TCSVT = {IEEE TCSVT})

@String(PR    = {PR})

@inproceedings{OneRestore,
  title={Onerestore: A universal restoration framework for composite degradation},
  author={Guo, Yu and Gao, Yuan and Lu, Yuxu and Zhu, Huilin and Liu, Ryan Wen and He, Shengfeng},
  booktitle={ECCV},
  pages={255--272},
  year={2024}
}

@inproceedings{clb_detection,
  title={Waterdiffusion: Learning a prior-involved unrolling diffusion for joint underwater saliency detection and visual restoration},
  author={Chang, Laibin and Wang, Yunke and Deng, Longxiang and Du, Bo and Xu, Chang},
  booktitle={AAAI},
  pages={1998--2006},
  year={2025}
}

@inproceedings{seg_1,
  title={UIS-Mamba: exploring mamba for underwater instance segmentation via dynamic tree scan and hidden state weaken},
  author={Cong, Runmin and Yu, Zongji and Fang, Hao and Sun, Haoyan and Kwong, Sam},
  booktitle={ACM MM},
  pages={343--352},
  year={2025}
}

@inproceedings{seg_2,
  title={Empowering dino representations for underwater instance segmentation via aligner and prompter},
  author={Chen, Zhiyang and Zhang, Chen and Fang, Hao and Cong, Runmin},
  booktitle={AAAI},
  pages={3201--3209},
  year={2026}
}

@article{3D_1,
  title={Underwater Scene Clarity Reconstruction via Multilayer Information Fusion and Self-Organized Stitching},
  author={Zhang, Weibo and Wang, Hao and Ren, Peng and Zhang, Weidong},
  journal={TCSVT},
  volume={36},
  number={2},
  pages={1848--1861},
  year={2026}
}

@inproceedings{3D_2,
  title={End-to-end underwater multi-view stereo for dense scene reconstruction},
  author={Yang, Guidong and Wen, Junjie and Zhao, Benyun and Li, Qingxiang and Huang, Yijun and Lei, Lei and Chen, Xi and Lam, Alan and Chen, Ben M},
  booktitle={ICRA},
  pages={7616--7623},
  year={2025}
}

@article{RRNet,
  title={{RRNet}: Relational reasoning network with parallel multiscale attention for salient object detection in optical remote sensing images},
  author={Cong, Runmin and Zhang, Yumo and Fang, Leyuan and Li, Jun and Zhao, Yao and Kwong, Sam},
  journal={TGRS},
  volume={60},
  pages={1--11},
  year={2021},
}

@article{Nested,
  title={Nested network with two-stream pyramid for salient object detection in optical remote sensing images},
  author={Li, Chongyi and Cong, Runmin and Hou, Junhui and Zhang, Sanyi and Qian, Yue and Kwong, Sam},
  journal={TGRS},
  volume={57},
  number={11},
  pages={9156--9166},
  year={2019}
}

@inproceedings{USUIR,
  title={Unsupervised underwater image restoration: From a homology perspective},
  author={Fu, Zhenqi and Lin, Huangxing and Yang, Yan and Chai, Shu and Sun, Liyan and Huang, Yue and Ding, Xinghao},
  booktitle={AAAI},
  volume={36},
  pages={643--651},
  year={2022}
}

@article{Gray_Edge,
  title={Edge-based color constancy},
  author={Van De Weijer, Joost and Gevers, Theo and Gijsenij, Arjan},
  journal={TIP},
  volume={16},
  number={9},
  pages={2207--2214},
  year={2007}
}

@inproceedings{CIDNet,
  title={Hvi: A new color space for low-light image enhancement},
  author={Yan, Qingsen and Feng, Yixu and Zhang, Cheng and Pang, Guansong and Shi, Kangbiao and Wu, Peng and Dong, Wei and Sun, Jinqiu and Zhang, Yanning},
  booktitle={CVPR},
  pages={5678--5687},
  year={2025}
}

@article{SSIM,
  title={Image quality assessment: from error visibility to structural similarity},
  author={Wang, Zhou and Bovik, Alan C and Sheikh, Hamid R and Simoncelli, Eero P},
  journal={TIP},
  volume={13},
  number={4},
  pages={600--612},
  year={2004}
}

@article{BaryIR,
  title={BaryIR: Learning multi-source unified representation in continuous barycenter space for generalizable all-in-one image restoration},
  author={Tang, Xiaole and He, Xiaoyi and Gu, Xiang and Sun, Jian},
  journal={arXiv preprint arXiv:2505.21637},
  year={2025}
}

@article{jaffe1990computer,
  title={Computer modeling and the design of optimal underwater imaging systems},
  author={Jaffe, Jules S},
  journal={JOE},
  volume={15},
  number={2},
  pages={101--111},
  year={1990}
}

@article{he2010single,
  title={Single image haze removal using dark channel prior},
  author={He, Kaiming and Sun, Jian and Tang, Xiaoou},
  journal={TPAMI},
  volume={33},
  number={12},
  pages={2341--2353},
  year={2010}
}

@incollection{zuiderveld1994contrast,
  title={Contrast limited adaptive histogram equalization},
  author={Zuiderveld, Karel},
  booktitle={Graphics Gems IV},
  pages={474--485},
  year={1994},
  publisher={Academic Press}
}

@inproceedings{ancuti2012enhancing,
  title={Enhancing underwater images and videos by fusion},
  author={Ancuti, Cosmin and Ancuti, Codruta Orniana and Haber, Tom and Bekaert, Philippe},
  booktitle={CVPR},
  pages={81--88},
  year={2012}
}

@article{zhang2022underwater,
  title={Underwater image enhancement via minimal color loss and locally adaptive contrast enhancement},
  author={Zhang, Weidong and Zhuang, Peixian and Sun, Hao-Hsiang and Li, Guohou and Kwong, Sam and Li, Chongyi},
  journal={TIP},
  volume={31},
  pages={3997--4010},
  year={2022}
}

@inproceedings{fu2022uncertainty,
  title={Uncertainty inspired underwater image enhancement},
  author={Fu, Zhenqi and Lin, Huangxing and Yang, Yan and Chai, Shu and Sun, Liyan and Huang, Yue and Ding, Xinghao},
  booktitle={ECCV},
  pages={465--482},
  year={2022}
}

@inproceedings{jiang2024five,
  title={Five A+ Network: You Only Need 9K Parameters for Underwater Image Enhancement},
  author={Jiang, Jingxia and Zheng, Tian and Wen, Jie and Zhang, Ting and Rigall, Eric and Jin, Sheng},
  booktitle={BMVC},
  year={2023},
  note={Paper 149}
}

@inproceedings{liang2021swinir,
  title={SwinIR: Image restoration using swin transformer},
  author={Liang, Jingyun and Cao, Jiezhang and Sun, Guolei and Zhang, Kai and Van Gool, Luc and Timofte, Radu},
  booktitle={ICCVW},
  pages={1833--1844},
  year={2021}
}

@inproceedings{chen2023dual,
  title={Dual Aggregation Transformer for Image Super-Resolution},
  author={Chen, Zheng and Zhang, Yulun and Gu, Jinjin and Kong, Linghe and Yang, Xiaokang and Yu, Fisher},
  booktitle={ICCV},
  pages={12312--12321},
  year={2023}
}

@inproceedings{wu2021contrastive,
  title={Contrastive learning for compact single image dehazing},
  author={Wu, Haiyan and Qu, Yanyun and Lin, Shaohui and Zhou, Jian and Qiao, Ruizhi and Zhang, Zhizhong and Xie, Yuan and Ma, Lizhuang},
  booktitle={CVPR},
  pages={10551--10560},
  year={2021}
}

@article{UniUIR,
  author={Zhang, Xu and Zhang, Huan and Wang, Guoli and Zhang, Qian and Zhang, Lefei and Du, Bo},
  title={UniUIR: Considering Underwater Image Restoration as an All-in-One Learner},
  journal={TIP},
  volume={34},
  pages={6963--6977},
  year={2025}
}

@inproceedings{MoCE-IR,
  title={Complexity experts are task-discriminative learners for any image restoration},
  author={Zamfir, Eduard and Wu, Zongwei and Mehta, Nancy and Tan, Yuedong and Paudel, Danda Pani and Zhang, Yulun and Timofte, Radu},
  booktitle={CVPR},
  pages={12753--12763},
  year={2025}
}

@inproceedings{UFO,
    author={Islam, Md Jahidul and Luo, Peigen and Sattar, Junaed},
    title={{Simultaneous Enhancement and Super-Resolution of Underwater Imagery 
    	    for Improved Visual Perception}},
    booktitle={RSS},
    year={2020},
    doi={{10.15607/RSS.2020.XVI.018}}
}

@inproceedings{Phaseformer,
  title={Phaseformer: Phase-based attention mechanism for underwater image restoration and beyond},
  author={Khan, Raqib and Negi, Anshul and Kulkarni, Ashutosh and Phutke, Shruti S and Vipparthi, Santosh Kumar and Murala, Subrahmanyam},
  booktitle={WACV},
  pages={9600--9611},
  year={2025}
}

@inproceedings{SMDR-IS,
  title={Synergistic multiscale detail refinement via intrinsic supervision for underwater image enhancement},
  author={Zhang, Dehuan and Zhou, Jingchun and Guo, Chunle and Zhang, Weishi and Li, Chongyi},
  booktitle={AAAI},
  pages={7033--7041},
  year={2024}
}

@article{CDF-UIE,
  title={CDF-UIE: Leveraging Cross-Domain Fusion for Underwater Image Enhancement},
  author={Zhang, Haopeng and Xu, Hongli and Yu, Xiaosheng and Zhang, Xiangyue and Gao, Xiujing and Wu, Chengdong},
  journal={TGRS},
  volume={63},
  pages={1--15},
  year={2025}
}

@inproceedings{WWE-UIE,
  title={WWE-UIE: A wavelet \& white balance efficient network for underwater image enhancement},
  author={Cheng, Ching-Heng and Lee, Jen-Wei and Lee, Chia-Ming and Hsu, Chih-Chung},
  booktitle={WACV},
  pages={2135--2145},
  year={2026}
}

@article{zzk_PR,
  title={Unveiling the underwater world: CLIP perception model-guided underwater image enhancement},
  author={Cao, Jiangzhong and Zeng, Zekai and Zhang, Xu and Zhang, Huan and Fan, Chunling and Jiang, Gangyi and Lin, Weisi},
  journal={PR},
  pages={111395},
  year={2025},
  publisher={Elsevier}
}

@inproceedings{AdaIR,
  title={AdaIR: Adaptive All-in-One Image Restoration via Frequency Mining and Modulation},
  author={Cui, Yuning and Zamir, Syed Waqas and Khan, Salman and Knoll, Alois and Shah, Mubarak and Khan, Fahad Shahbaz},
  booktitle={ICLR},
  year={2025}
}

@ARTICLE{FSMC,
  author={Qiao, Nianzu and Sun, Jia and Ge, Quanbo and Sun, Changyin},
  journal={TCSVT},
  title={UIE-FSMC: Underwater Image Enhancement Based on Few-Shot Learning and Multi-Color Space}, 
  year={2023},
  volume={33},
  number={10},
  pages={5391--5405}}

@ARTICLE{WWVP,
  author={Zhang, Weidong and Zhou, Ling and Zhuang, Peixian and Li, Guohou and Pan, Xipeng and Zhao, Wenyi and Li, Chongyi},
  journal={TCSVT},
  title={Underwater Image Enhancement via Weighted Wavelet Visual Perception Fusion}, 
  year={2024},
  volume={34},
  number={4},
  pages={2469--2483}}

@inproceedings{MPerceiver,
  title={Multimodal Prompt Perceiver: Empower Adaptiveness Generalizability and Fidelity for All-in-One Image Restoration},
  author={Ai, Yuang and Huang, Huaibo and Zhou, Xiaoqiang and Wang, Jiexiang and He, Ran},
  booktitle={CVPR},
  pages={25432--25444},
  year={2024}
}

@inproceedings{GUPDM,
  title={A generalized physical-knowledge-guided dynamic model for underwater image enhancement},
  author={Mu, Pan and Xu, Hanning and Liu, Zheyuan and Wang, Zheng and Chan, Sixian and Bai, Cong},
  booktitle={ACM MM},
  pages={7111--7120},
  year={2023}
}

@ARTICLE{zhang_TCSVT_PCFB,
  author={Zhang, Weidong and Liu, Qingmin and Feng, Yikun and Cai, Lei and Zhuang, Peixian},
  journal={TCSVT},
  title={Underwater Image Enhancement via Principal Component Fusion of Foreground and Background}, 
  year={2024},
  volume={34},
  number={11},
  pages={10930--10943}}

@article{zhou_TCSVT_FDCE,
  title={FDCE-Net: Underwater Image Enhancement with Embedding Frequency and Dual Color Encoder},
  author={Cheng, Zheng and Fan, Guodong and Zhou, Jingchun and Gan, Min and Chen, C. L. Philip},
  journal={TCSVT},
  volume={35},
  number={2},
  pages={1728--1744},
  year={2025}
}

@article{zhou_IJCV_HCLR,
  title={HCLR-net: Hybrid contrastive learning regularization with locally randomized perturbation for underwater image enhancement},
  author={Zhou, Jingchun and Sun, Jiaming and Li, Chongyi and Jiang, Qiuping and Zhou, Man and Lam, Kin-Man and Zhang, Weishi and Fu, Xianping},
  journal={IJCV},
  volume={42},
  pages={4132--4156},
  year={2024},
}

@article{zhou_IJCV_ADPP,
  title={Underwater image restoration via adaptive dark pixel prior and color correction},
  author={Zhou, Jingchun and Liu, Qian and Jiang, Qiuping and Ren, Wenqi and Lam, Kin-Man and Zhang, Weishi},
  journal={IJCV},
  pages={1--19},
  year={2023}
}

@inproceedings{Semi-UIR,
  title={Contrastive semi-supervised learning for underwater image restoration via reliable bank},
  author={Huang, Shirui and Wang, Keyan and Liu, Huan and Chen, Jun and Li, Yunsong},
  booktitle={CVPR},
  pages={18145--18155},
  year={2023}
}

@article{EUVP,
  title={Fast underwater image enhancement for improved visual perception},
  author={Islam, Md Jahidul and Xia, Youya and Sattar, Junaed},
  journal={RA-L},
  volume={5},
  number={2},
  pages={3227--3234},
  year={2020},
  publisher={IEEE}
}

@article{UCCS,
  title={Real-world underwater enhancement: Challenges, benchmarks, and solutions under natural light},
  author={Liu, Risheng and Fan, Xin and Zhu, Ming and Hou, Minjun and Luo, Zhongxuan},
  journal={TCSVT},
  volume={30},
  number={12},
  pages={4861--4875},
  year={2020},
  publisher={IEEE}
}

@article{2,
  title={A Survey on Underwater Image Enhancement Techniques},
  author={Sahu, Pooja and Gupta, Neelesh and Sharma, Neetu},
  journal={IJCA},
  volume={87},
  number={13},
  pages={19--23},
  year={2014}
}

@book{3,
  title={Marine ecology: processes, systems, and impacts},
  author={Kaiser, Michel J and others},
  year={2011},
  publisher={Oxford University Press, USA}
}

@article{Ucolor,
  title={Underwater image enhancement via medium transmission-guided multi-color space embedding},
  author={Li, Chongyi and Anwar, Saeed and Hou, Junhui and Cong, Runmin and Guo, Chunle and Ren, Wenqi},
  journal={TIP},
  volume={30},
  pages={4985--5000},
  year={2021},
  publisher={IEEE}
}

@article{Ushape,
  author={Peng, Lintao and Zhu, Chunli and Bian, Liheng},
  title={U-Shape Transformer for Underwater Image Enhancement},
  journal={TIP},
  volume={32},
  pages={3066--3079},
  year={2023}
}

@inproceedings{NU2Net,
  title={Underwater ranker: Learn which is better and how to be better},
  author={Guo, Chunle and Wu, Ruiqi and Jin, Xin and Han, Linghao and Zhang, Weidong and Chai, Zhi and Li, Chongyi},
  booktitle={AAAI},
  pages={702--709},
  year={2023}
}

@article{IBLA,
  title={Underwater image restoration based on image blurriness and light absorption},
  author={Peng, Yan-Tsung and Cosman, Pamela C},
  journal={TIP},
  volume={26},
  number={4},
  pages={1579--1594},
  year={2017},
  publisher={IEEE}
}

@article{U45,
  title={A fusion adversarial underwater image enhancement network with a public test dataset},
  author={Li, H and Li, J and Wang, W},
  journal={arXiv preprint arXiv:1906.06819},
  year={2019}
}

@inproceedings{UDCP,
  title={Transmission estimation in underwater single images},
  author={Drews, Paul and Nascimento, Erickson and Moraes, Filipe and Botelho, Silvia and Campos, Mario},
  booktitle={ICCVW},
  pages={825--830},
  year={2013}
}

@article{MLLE,
  title={Underwater image enhancement via minimal color loss and locally adaptive contrast enhancement},
  author={Zhang, Weidong and Zhuang, Peixian and Sun, Hai-Han and Li, Guohou and Kwong, Sam and Li, Chongyi},
  journal={TIP},
  volume={31},
  pages={3997--4010},
  year={2022},
  publisher={IEEE}
}

@article{FUnIE,
  title={Fast underwater image enhancement for improved visual perception},
  author={Islam, Md Jahidul and Xia, Youya and Sattar, Junaed},
  journal={RA-L},
  volume={5},
  number={2},
  pages={3227--3234},
  year={2020},
  publisher={IEEE}
}

@article{UIQM,
  title={Human-visual-system-inspired underwater image quality measures},
  author={Panetta, Karen and Gao, Chen and Agaian, Sos},
  journal={JOE},
  volume={41},
  number={3},
  pages={541--551},
  year={2015}
}

@article{UCIQE,
  title={An underwater color image quality evaluation metric},
  author={Yang, Miao and Sowmya, Arcot},
  journal={TIP},
  volume={24},
  number={12},
  pages={6062--6071},
  year={2015}
}

@InProceedings{LPIPS,
author = {Zhang, Richard and Isola, Phillip and Efros, Alexei A. and Shechtman, Eli and Wang, Oliver},
title = {The Unreasonable Effectiveness of Deep Features as a Perceptual Metric},
booktitle={CVPR},
pages={586--595},
year = {2018}
}

@ARTICLE{UIEB,
  author={Li, Chongyi and Guo, Chunle and Ren, Wenqi and Cong, Runmin and Hou, Junhui and Kwong, Sam and Tao, Dacheng},
  journal={TIP},
  title={An Underwater Image Enhancement Benchmark Dataset and Beyond}, 
  year={2020},
  volume={29},
  number={},
  pages={4376--4389}}

@InProceedings{DRSformer,
    author={Chen, Xiang and Li, Hao and Li, Mingqiang and Pan, Jinshan}, 
    title={Learning a Sparse Transformer Network for Effective Image Deraining},
    booktitle={CVPR},
    month={June},
    year={2023},
    pages={5896--5905}
}

@inproceedings{VGG,
  title={Very Deep Convolutional Networks for Large-Scale Image Recognition},
  author={Simonyan, Karen and Zisserman, Andrew},
  booktitle={ICLR},
  year={2015}
}

@inproceedings{Restormer,
  title={Restormer: Efficient transformer for high-resolution image restoration},
  author={Zamir, Syed Waqas and Arora, Aditya and Khan, Salman and Hayat, Munawar and Khan, Fahad Shahbaz and Yang, Ming-Hsuan},
  booktitle={CVPR},
  pages={5728--5739},
  year={2022}
}

@inproceedings{AirNet,
  title={All-in-one image restoration for unknown corruption},
  author={Li, Boyun and Liu, Xiao and Hu, Peng and Wu, Zhongqin and Lv, Jiancheng and Peng, Xi},
  booktitle={CVPR},
  pages={17452--17462},
  year={2022}
}

@inproceedings{PromptIR,
  title={PromptIR: Prompting for All-in-One Blind Image Restoration},
  author={Potlapalli, Vaishnav and Zamir, Syed Waqas and Khan, Salman and Khan, Fahad Shahbaz},
  booktitle={NeurIPS},
  pages={71275--71293},
  year={2023}
}

\clearpage
\appendix

\twocolumn[
\begin{center}
    {\LARGE\bfseries Supplementary Material\par}
    \vspace{0.35em}


\end{center}

\vspace{0.8em}
]

\section*{Appendix}

\setcounter{table}{0}
\renewcommand{\thetable}{S\arabic{table}}

\setcounter{figure}{0}
\renewcommand{\thefigure}{S\arabic{figure}}



\subsection{Organisation}
We first provide the dataset and evaluation protocols, followed by additional
qualitative results, representative failure cases, and loss ablations. We then
analyse the two control paths:
spatial guidance through same-checkpoint, per-image, cross-dataset, and
real-degradation analyses; and latent control through loss-weight,
representation, and gate diagnostics.

\subsection{Additional Experimental Details}

\noindent\textbf{Dataset configuration.}

\begin{table}[!ht]
\centering
\renewcommand{\arraystretch}{1.1}

\resizebox{\linewidth}{!}{
\begin{tabular}{c|c|c|c|c}

\toprule
\textbf{Setting} & \textbf{Train/Test} & \textbf{Dataset} & \textbf{Type} & \textbf{No. of Images}
\\

\midrule

\multirow{6}{*}{Set \textbf{\textcircled{1}}}
& \multirow{1}{*}{Train}
    & UIEB   & Ref.   & 800
    \\
    \cline{2-5}

  & \multirow{5}{*}{Test} & U90      & Ref.   & 90
  \\
  & & Challenge-60      & Non-ref. & 60  \\
  & & U45        & Non-ref. & 45  \\
  & & UCCS             & Non-ref. & 300 \\
  & & EUVP-330          & Non-ref. & 330 \\

\midrule
\multirow{2}{*}{Set \textbf{\textcircled{2}}}
  & Train & LSUI   & Ref.   & 3,879 \\
  & Test  & LSUI & Ref.   & 400  \\

\midrule
\multirow{3}{*}{Set \textbf{\textcircled{3}}}
  & Train & UFO   & Ref.   & 1,200 \\
  & val   & UFO   & Ref.   & 300 \\
  & Test  & UFO   & Ref.   & 120  \\

\midrule
  \multirow{3}{*}{Set \textbf{\textcircled{4}}}
  & Train & EUVP-Scene   & Ref.   & 1,748\\
  & val   & EUVP-Scene   & Ref.   & 218 \\
  & Test  & EUVP-Scene   & Ref.   & 219  \\

\midrule
  \multirow{3}{*}{Set \textbf{\textcircled{5}}}
  & Train & EUVP-Dark   & Ref.   & 4,440\\
  & val   & EUVP-Dark   & Ref.   & 555 \\
  & Test  & EUVP-Dark   & Ref.   & 555  \\

\bottomrule
\end{tabular}}
\caption{Dataset configuration for training and testing.}
\label{tab:dataset}
\end{table}

\noindent\textbf{Evaluation protocol and result provenance.}
The benchmark tables combine results reported by the cited methods with
PROTEUS results evaluated using our checkpoints. They should therefore be
viewed as benchmark-level comparisons rather than jointly retrained results
under a unified codebase. For paired evaluation, each PROTEUS model uses the
dataset-specific checkpoint and held-out split listed in
Table~\ref{tab:dataset}. Inputs and references are resized to
$256\times256$, following the validation protocol used for checkpoint
selection, and LPIPS is computed with the AlexNet backbone. Non-reference
results are obtained using the UIEB-trained checkpoint at the same resolution.
Parameter counts and FLOPs are computed with a $3\times256\times256$ input.

\subsection{Additional Qualitative Results}

\begin{figure*}[!t]
\centering
\includegraphics[page=1,width=0.9\textwidth]{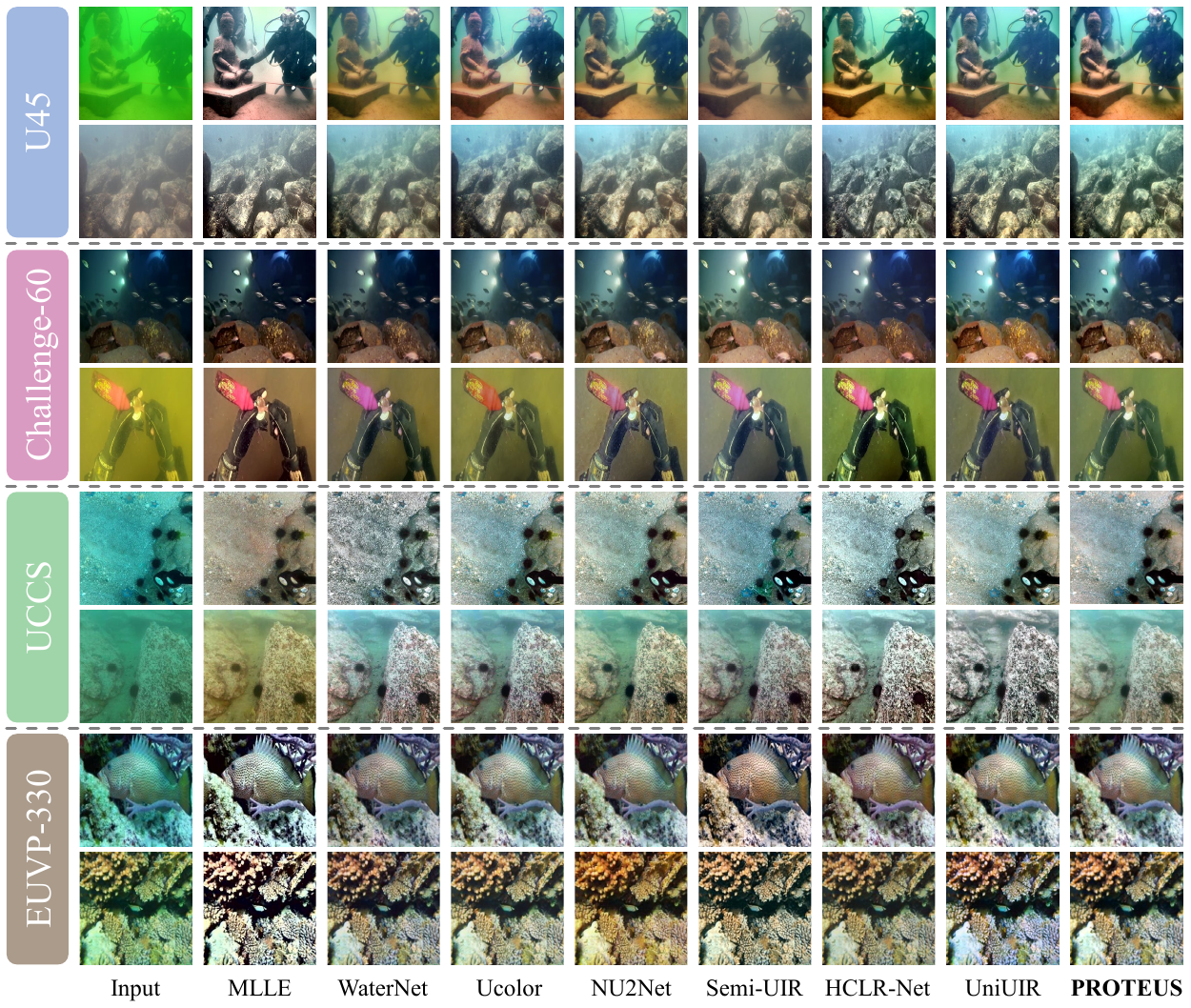}
\caption{Visual comparisons on U45~\cite{U45}, Challenge-60~\cite{UIEB}, UCCS~\cite{UCCS}, and EUVP-330~\cite{EUVP}.}
\label{fig:non_ref}
\end{figure*}

On paired scenes, PROTEUS corrects severe green, blue, and yellow colour casts
while preserving natural white points and structural detail. In low-light
scenes, it restores luminance without the noise amplification or over-bluing
observed in several competing methods. On non-reference data, PROTEUS removes
strong colour casts while preserving skin tones, equipment colours, and coral
textures, while maintaining spatially coherent enhancement across diverse scenes.

\begin{figure}[!t]
\centering
\includegraphics[width=1\columnwidth]{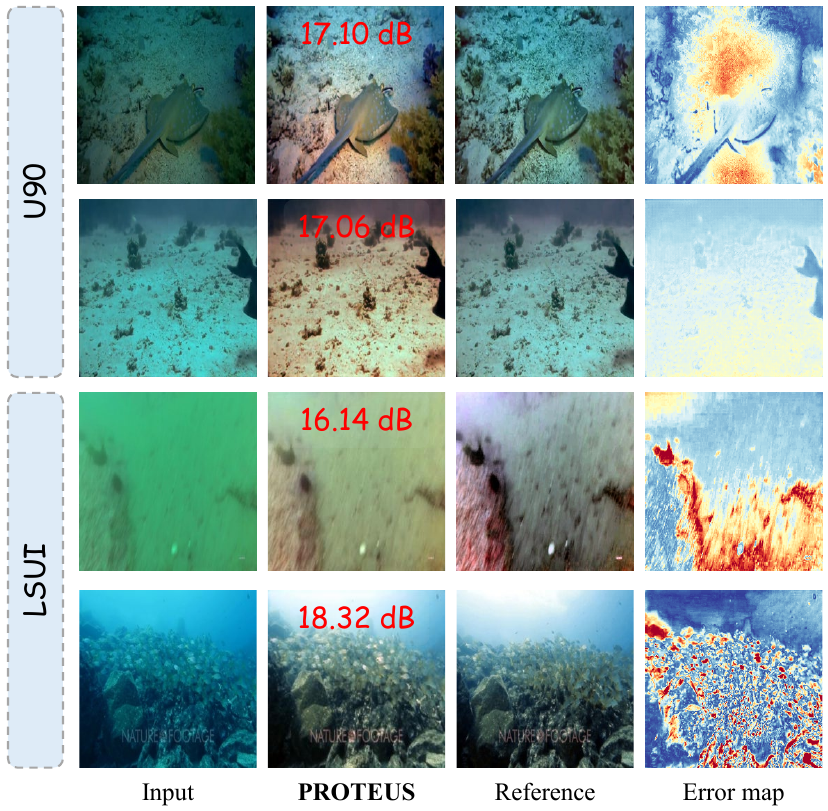}
\caption{{Representative low-PSNR cases from the paired U90 and
LSUI test sets. All panels follow the $256\times256$ evaluation protocol.
Error maps show the per-pixel mean absolute RGB difference between PROTEUS and
the reference under one shared colour scale. These examples are selected from
the low-performance tail to expose limitations and do not estimate their
frequency over the full test sets.}}
\label{fig:failure_cases}
\end{figure}

The complete non-reference comparison in Fig.~\ref{fig:non_ref} complements
the paired examples in the main paper. Figure~\ref{fig:failure_cases} shows
the opposite end of the performance distribution. PROTEUS can over-correct
colour or illumination when the reference remains close to the underwater
input, while severely blurred or low-contrast inputs expose information that
cannot be reliably reconstructed. Paired references also encode one target
colour and exposure, so pixelwise metrics may penalise visually plausible but
differently balanced outputs. These cases motivate confidence-aware correction
and objectives that distinguish irreversible information loss from correctable
colour degradation.


\subsection{Additional Ablation Results}

Component-level ablations are reported and discussed in the main paper. Here
we complement them with the loss ablation in Table~\ref{tab:ablation_loss}; all
absolute quality metrics use the same precision as the main tables.


\begin{table}[!ht]
\centering
\begin{tabular}{l|ccc}
\toprule
Setting & PSNR$\uparrow$ & SSIM$\uparrow$ & LPIPS$\downarrow$ \\
\midrule
w/o $\mathcal{L}_{{center}}$ & 25.24 & 0.927 & 0.098 \\
w/o $\mathcal{L}_{{ssl}}$  & 25.31 & 0.928 & 0.095 \\
w/o $\mathcal{L}_{{contrast}}$  & \underline{25.39} & \underline{0.929} & \underline{0.092} \\
w/o $\mathcal{L}_{{orth}}$  & 25.32 & \underline{0.929} & 0.094 \\
\textbf{All loss} & \textbf{25.50} & \textbf{0.933} & \textbf{0.081} \\
\bottomrule
\end{tabular}
\caption{{Ablation study on loss components. The
full-objective row is aligned with the final checkpoint reported in the
main paper; loss-removal rows are independently trained
runs rather than seed-paired test-time interventions.}}
\label{tab:ablation_loss}
\end{table}

Removing $\mathcal{L}_{center}$ causes the largest degradation, reducing PSNR
from 25.50 to 25.24 and increasing LPIPS from 0.081 to 0.098. Excluding
$\mathcal{L}_{ssl}$ also degrades performance, while removing
$\mathcal{L}_{contrast}$ or $\mathcal{L}_{orth}$ produces smaller but
consistent declines. The full objective achieves the best overall results.
Since the reduced-objective variants come from independent training runs, the
differences are regarded as descriptive ablation evidence rather than paired
statistical effects.


\begin{table*}[!t]
\centering
\setlength{\tabcolsep}{5.0pt}
\begin{tabular}{l|ccc|ccc}
\toprule
\multirow{2}{*}{\textbf{Guide intervention}} &
\multicolumn{3}{c|}{\textbf{U90}} &
\multicolumn{3}{c}{\textbf{LSUI}} \\
\cmidrule(lr){2-4}\cmidrule(lr){5-7}
& PSNR$\uparrow$ & SSIM$\uparrow$ & LPIPS$\downarrow$
& PSNR$\uparrow$ & SSIM$\uparrow$ & LPIPS$\downarrow$ \\
\midrule
Zero guide
& 22.06 & 0.890 & 0.134
& 24.70 & 0.889 & 0.152 \\
Uniform guide
& 24.58 & 0.927 & 0.086
& 24.49 & 0.881 & 0.161 \\
Raw-input guide
& \underline{25.47} & \textbf{0.936} & \textbf{0.080}
& \underline{28.85} & \underline{0.921} & \underline{0.085} \\
Cross-image guide
& 23.76 & 0.918 & 0.103
& 27.58 & 0.911 & 0.099 \\
Full spatial guide
& \textbf{25.50} & \underline{0.935} & \underline{0.081}
& \textbf{28.99} & \textbf{0.921} & \textbf{0.083} \\
\bottomrule
\end{tabular}
\caption{{Same-checkpoint analysis of spatial guidance on all
90 U90 pairs and 400 LSUI pairs. ``Uniform'' broadcasts the per-channel spatial
mean, ``raw-input'' bypasses guide preprocessing while preserving the normal
main path, and ``cross-image'' uses the guide from the next image in
deterministic dataset order. All model weights remain fixed; absolute metrics
follow the PSNR/SSIM/LPIPS precision used in the main paper, while best and
second-best markings follow the unrounded measurements.}}
\label{tab:spatial_guidance}
\end{table*}

\subsection{Spatial-Guidance Analysis}

\subsubsection{Same-checkpoint guide interventions.}
{Table~\ref{tab:spatial_guidance} separates the presence,
spatial variation, preprocessing, and image correspondence of the guide
without confounding them with retraining. Relative to the full model, the
uniform guide reduces PSNR by 0.925\,dB on U90 (95\% paired bootstrap CI
$[0.425,1.450]$) and 4.510\,dB on LSUI
($[4.086,4.928]$). Supplying a guide from another image also causes clear
drops of 1.741\,dB ($[1.063,2.459]$) and 1.415\,dB
($[1.118,1.731]$), respectively. Thus, the gain is not explained by adding an
arbitrary auxiliary image or a global colour statistic; spatially aligned
guidance is important.}

The raw underwater input is nearly as effective as the
preprocessed guide: the full-minus-raw PSNR difference is 0.036\,dB on U90
($[-0.076,0.145]$) and 0.147\,dB on LSUI
($[0.095,0.202]$). It also gives slightly better SSIM and LPIPS on U90.
Accordingly, this intervention supports the need for spatial correspondence
more strongly than a claim that one particular guide preprocessing operation
is universally optimal. Together with the real-data stratification below, the
result positions GDFMB as spatially aligned degradation conditioning rather
than generic global degradation embedding.

\begin{figure*}[!t]
\centering
\includegraphics[width=0.9\textwidth]{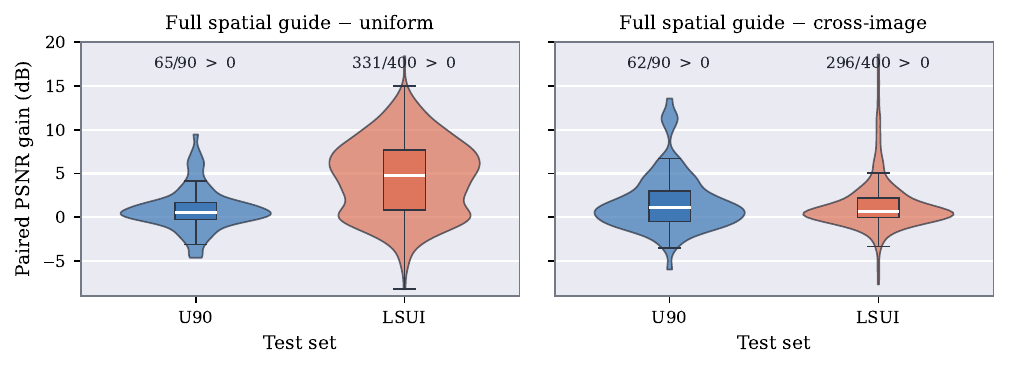}
\caption{{Per-image paired PSNR gains of the full spatial guide
over uniform and cross-image guide interventions. Boxes show the interquartile
range and median; violins show the complete empirical distribution. The
fractions give the number of images with positive gain. Although individual
counterexamples remain, the full guide improves most images on both test
sets, showing that the mean gains in Table~\ref{tab:spatial_guidance} are not
driven by only a few outliers.}}
\label{fig:spatial_guidance_distribution}
\end{figure*}

\subsubsection{Per-Image Gains.}

{Figure~\ref{fig:spatial_guidance_distribution} complements the
mean and confidence-interval analysis with paired image-level distributions.
The full guide outperforms the uniform intervention on 65/90 U90 and 331/400
LSUI images, with median gains of 0.537 and 4.730\,dB, respectively. Against
the cross-image intervention, it improves 62/90 and 296/400 images, with
median gains of 1.154 and 0.671\,dB. The negative tails show that spatial
guidance is not uniformly optimal for every individual pair; nevertheless,
the positive medians and majority counts on both datasets make the aggregate
effect more robust than a mean-only comparison.}

\begin{table*}[!t]
\centering
\setlength{\tabcolsep}{3.2pt}
{%
\begin{tabular}{ll|ccc}
\toprule
\textbf{Target set} & \textbf{Intervention} &
$\Delta$PSNR [95\% CI]$\uparrow$ &
$\Delta$SSIM [95\% CI]$\uparrow$ &
$\Delta$LPIPS [95\% CI]$\uparrow$ \\
\midrule
\multirow{3}{*}{LSUI}
& Uniform guide
& $0.29\,[0.13,0.44]$ & $0.007\,[0.005,0.010]$ & $-0.003\,[-0.005,-0.001]$ \\
& Identity (all-one) gate
& $1.25\,[1.00,1.48]$ & $0.028\,[0.024,0.031]$ & $0.027\,[0.022,0.032]$ \\
& Both reduced
& $2.03\,[1.71,2.36]$ & $0.045\,[0.040,0.050]$ & $0.037\,[0.031,0.043]$ \\
\midrule
\multirow{3}{*}{UFO-120}
& Uniform guide
& $0.39\,[0.08,0.72]$ & $0.009\,[0.003,0.016]$ & $-0.001\,[-0.006,0.005]$ \\
& Identity (all-one) gate
& $1.41\,[1.07,1.76]$ & $0.033\,[0.028,0.038]$ & $0.030\,[0.024,0.036]$ \\
& Both reduced
& $2.13\,[1.64,2.65]$ & $0.058\,[0.049,0.066]$ & $0.041\,[0.031,0.051]$ \\
\midrule
\multirow{3}{*}{EUVP-Scene}
& Uniform guide
& $0.41\,[0.21,0.60]$ & $0.016\,[0.012,0.020]$ & $0.002\,[-0.001,0.005]$ \\
& Identity (all-one) gate
& $1.60\,[1.40,1.81]$ & $0.040\,[0.035,0.045]$ & $0.043\,[0.037,0.048]$ \\
& Both reduced
& $2.20\,[1.89,2.49]$ & $0.064\,[0.057,0.070]$ & $0.049\,[0.042,0.055]$ \\
\midrule
\multirow{3}{*}{EUVP-Dark}
& Uniform guide
& $0.01\,[-0.05,0.08]$ & $0.009\,[0.007,0.011]$ & $-0.005\,[-0.006,-0.003]$ \\
& Identity (all-one) gate
& $0.78\,[0.63,0.91]$ & $0.029\,[0.026,0.032]$ & $0.013\,[0.010,0.016]$ \\
& Both reduced
& $0.96\,[0.80,1.13]$ & $0.043\,[0.040,0.047]$ & $0.011\,[0.008,0.015]$ \\
\bottomrule
\end{tabular}}
\caption{{Zero-shot transfer of the two controls. The
UIEB-trained checkpoint is evaluated without fine-tuning on four disjoint
paired target datasets. Uniform guide, identity (all-one) gate, and both reduced are
same-checkpoint inference interventions. For PSNR/SSIM, $\Delta$ is full minus
intervention; for LPIPS, it is intervention minus full, so positive values
always favour the complete control. Intervals use 10,000 image-level paired
bootstrap samples. The displayed precision follows the main-paper convention:
two decimals for PSNR and three for SSIM/LPIPS.}}
\label{tab:cross_dataset_control}
\end{table*}

\subsubsection{Cross-Dataset Control Transfer.}

{Table~\ref{tab:cross_dataset_control} tests whether the frozen
UIEB model continues to use both controls after a distribution shift. The
identity (all-one) gate intervention degrades all three metrics on every target set,
with PSNR reductions from 0.776 to 1.602\,dB. Reducing both controls gives
still larger PSNR drops of 0.965--2.198\,dB. Spatial variation alone has a
more metric-dependent effect: the full guide improves PSNR and SSIM
significantly on LSUI, UFO-120, and EUVP-Scene, but its EUVP-Dark PSNR
interval includes zero; LPIPS slightly favours the uniform guide on LSUI and
EUVP-Dark. Thus, the test supports cross-dataset functional use of the latent
skip modulation and broadly supports spatial guidance for fidelity, without
claiming that every control improves every perceptual metric under arbitrary
domain shift.}

\begin{table*}[!t]
\centering
\small
\setlength\tabcolsep{2.8pt}
\begin{tabular}{lccc|cc}
\toprule
\textbf{Setting} & $\lambda_a$ & $\lambda_o$ & $\lambda_c$
& $\Delta$MSE$\uparrow$ & $\Delta$Cos$\uparrow$ \\
\midrule
Baseline       & 0.10 & 0.05 & 0.05 & $-16.23{\pm}0.56\%$ & $-0.1048{\pm}0.0017$ \\
Align-heavy    & 0.16 & 0.02 & 0.02 & $-12.26{\pm}0.37\%$ & $-0.0927{\pm}0.0011$ \\
Align-dominant & 0.18 & 0.01 & 0.01 & \underline{$-10.58{\pm}0.32\%$} & \underline{$-0.0876{\pm}0.0009$} \\
Align-only     & 0.20 & 0.00 & 0.00 & \textbf{$-7.42{\pm}0.37\%$} & \textbf{$-0.0779{\pm}0.0011$} \\
Orth-heavy     & 0.02 & 0.16 & 0.02 & $-18.93{\pm}0.72\%$ & $-0.1131{\pm}0.0023$ \\
Contrast-heavy & 0.02 & 0.02 & 0.16 & $-22.50{\pm}0.60\%$ & $-0.1244{\pm}0.0019$ \\
\bottomrule
\end{tabular}
\caption{Fixed-coordinate local sensitivity to the relative latent-loss
weights. All settings share the same initial checkpoint, frozen
encoder--decoder, five-epoch adaptation budget, and U90 evaluation protocol.
The reported columns are seed-paired representation changes from the
same-budget baseline, not absolute restoration scores; values are mean $\pm$
sample standard deviation over three seeds.}
\label{tab:lambda_scan}
\end{table*}

\subsubsection{Real Spatial-Degradation Stratification.}
To avoid relying on visually implausible synthetic perturbations, we analyse
only the original aligned input--reference pairs. Let
$\mathbf{d}(x)=\operatorname{Lab}(\mathbf{I}_{in}(x))-
\operatorname{Lab}(\mathbf{I}_{gt}(x))$. We average this residual over an
$8\times8$ grid and remove its image-wide mean so that a uniform global colour
cast is not counted as spatial non-uniformity. The score is
\begin{equation}
s(\mathbf{I})=
\sqrt{\frac{1}{64}\sum_{b=1}^{64}
\left\|\overline{\mathbf{d}}_{b}-\overline{\mathbf{d}}\right\|_2^2}.
\end{equation}
The score is fixed before model inference and does not use any restoration
output. Stable rank splitting partitions each dataset into five balanced score
intervals, containing 18 U90 and 80 LSUI images per interval. These ordered
groups quantify relative spatial non-uniformity rather than a calibrated scale
of absolute physical degradation severity.

\begin{figure*}[!t]
\centering
\includegraphics[width=0.9\textwidth]{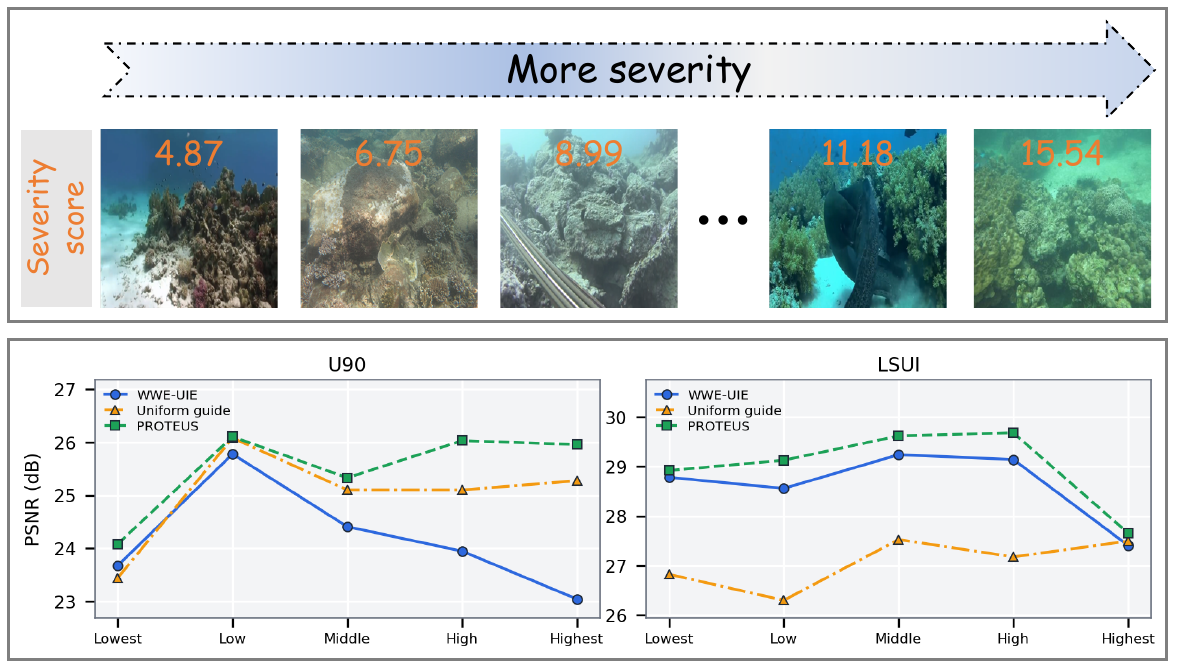}
\caption{Real spatial-degradation analysis on the paired U90 and
LSUI test sets. Samples are ranked by a model-independent low-frequency
CIELAB residual score computed from the original input--reference pairs and
split into five balanced within-dataset intervals. The top row shows
representative U90 inputs ordered by this relative spatial non-uniformity
score, rather than a calibrated measure of absolute physical degradation
severity. ``Uniform guide'' replaces each spatial guide with its per-channel
spatial mean while all network weights remain fixed. PROTEUS achieves the
highest mean PSNR in all ten score intervals, whereas the relative gain across
intervals is dataset dependent.}
\label{fig:real_spatial_severity}
\end{figure*}

\begin{figure*}[!t]
\centering
\includegraphics[width=1\textwidth]{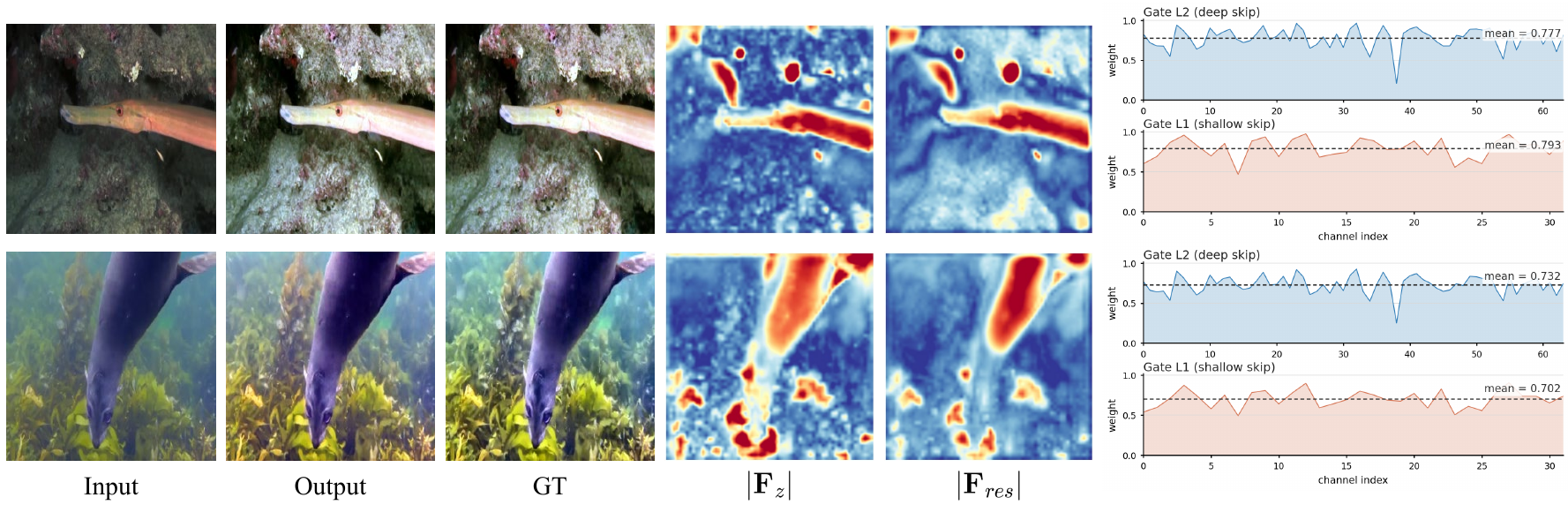}
\caption{Representation diagnostics on U90. The visualization shows input,
restored output, ground truth, control-code and complementary-residual
magnitudes, and channel-wise gate weights.}
\label{fig:latent_gate}
\end{figure*}

As shown in Fig.~\ref{fig:real_spatial_severity}, PROTEUS achieves higher mean
PSNR than the uniform-guide intervention in every score interval. From the
lowest to the highest score interval, the mean PSNR gains are
$0.640/0.024/0.230/0.931/0.683$\,dB on U90 and
$2.095/2.829/2.095/2.504/0.160$\,dB on LSUI. The variation across the ordered
intervals is dataset dependent. Retaining spatially varying guidance is thus
beneficial on average across real degradation strata, but the result does not
support a universal claim that the benefit must grow monotonically with
degradation severity.

\subsection{Latent-Control Analysis}

\subsubsection{Latent-Loss Weight Sensitivity.}
We test whether changing the relative latent-loss weights makes
$\mathbf{F}_z$ closer than the unfiltered bottleneck feature to a separately
encoded ground-truth latent. This is a fixed-coordinate local sensitivity
study, not a from-scratch component ablation. Starting from the reported
checkpoint, we freeze the encoder and decoder and adapt only the mapper and two
Attention Gates (113,296 parameters) for five epochs on the 800-pair UIEB
training split. We use seeds 7, 17, and 27 and keep
$\lambda_a+\lambda_o+\lambda_c=0.20$. All settings are evaluated at epoch 5
on the disjoint 90-pair U90 split.

Because this protocol adds an adaptation stage, it is not a restoration
ablation, and its adapted scores are not comparable to the unadapted
25.50\,dB main-paper result. We therefore omit PSNR/SSIM and use
Table~\ref{tab:lambda_scan} only for latent-loss sensitivity diagnostics.
Here, $\Delta$MSE is
$(\mathrm{MSE}_{raw}-\mathrm{MSE}_{F_z})/\mathrm{MSE}_{raw}$, and
$\Delta$Cos is
$\cos(\mathbf{F}_z,\mathbf{F}_{gt,z})-\cos(\mathbf{F}_l,\mathbf{F}_{gt,z})$;
positive values indicate improvement over the raw bottleneck.

Increasing $\lambda_a$ reduces the MSE deficit from $16.23\%$ to $7.42\%$
and improves the cosine gap from $-0.1048$ to $-0.0779$. Conversely,
orthogonality- and contrast-heavy controls enlarge the MSE deficit to
$18.93\%$ and $22.50\%$. No setting crosses zero on either diagnostic.
Relative-weight calibration
therefore does not justify a clean-latent projection claim; it instead
supports the latent objectives as optimisation biases for a task-oriented
control code.

\subsubsection{Representation Diagnostics.}

We evaluate 90 U90 pairs using the latent codes and gate weights. The mean
absolute cosine similarity between $\mathbf{F}_{{z}}$ and
$\mathbf{F}_{{res}}$ is 0.0077, and the mean gate activation is 0.7657,
corresponding to 23.4\% average attenuation of skip features. Direct proximity
to a separately encoded ground-truth latent is not monotonically improved: the
unfiltered bottleneck latent has MSE 0.0145 and cosine similarity 0.8067,
whereas $\mathbf{F}_{{z}}$ has MSE 0.0164 and cosine similarity 0.7162. These
results do not support interpreting $\mathbf{F}_{{z}}$ as a metrically cleaner
embedding. Instead, its complementary direction and measurable gate response
support its operational role as a task-oriented controller for skip-feature
modulation.


\subsubsection{Gate Mechanism Diagnostics.}
The gate interventions reported in Table~5 of the main paper use the same final
checkpoint with all network weights fixed. The fixed gate is estimated from
the disjoint UIEB training split, while channel-shuffled and cross-image
results are averaged over five deterministic repeats per image. Relative to
the predicted gate, the all-one and channel-shuffled settings reduce PSNR by
3.35 and 1.07\,dB, respectively. Fixed training-mean and cross-image gates
reduce PSNR by 0.27 and 0.42\,dB. These same-checkpoint interventions indicate
that learned attenuation, channel assignment, and input-conditioned gate
prediction all contribute to restoration. Because the intervention results do
not localise degradation within individual channels, we retain the narrower
interpretation of latent-conditioned skip modulation.

We additionally compare each encoder channel's normalized degraded/ground-truth
feature discrepancy with its suppression, $1-g$. At the deeper gate, the mean
Spearman correlation is $0.065$ (95\% CI $[0.044,0.086]$), and the most
sensitive channel quartile receives lower activation than the least sensitive
quartile (0.744 vs.\ 0.757). At the shallower gate, however, the direction
reverses: the correlation is $-0.104$ ($[-0.139,-0.070]$), with activations
0.761 vs.\ 0.733. These scale-dependent results do not justify treating either
the gate or $\mathbf{F}_z$ as a direct detector of degradation-contaminated
channels. We therefore use the narrower interpretation of latent-conditioned
skip modulation.

\end{document}